\documentclass[conference]{IEEEtran}
\IEEEoverridecommandlockouts
\usepackage[english]{babel}
\usepackage[utf8]{inputenc}
\usepackage{cite}
\usepackage{amsmath,amssymb,amsfonts}
\usepackage{algorithmic}
\usepackage{graphicx}
\usepackage{textcomp}
\usepackage{xcolor}
\usepackage[utf8]{inputenc}
\usepackage{subfigure}

\def\BibTeX{{\rm B\kern-.05em{\sc i\kern-.025em b}\kern-.08emT\kern-.1667em\lower.7ex\hbox{E}\kern-.125emX}}

\begin{document}
	
	\title{On the use of local structural properties for improving the efficiency of hierarchical community detection methods}
	\author{
		\IEEEauthorblockN{Julio-Omar Palacio-Niño}
		\IEEEauthorblockA{
			\textit{Dept. Computer Science and Artificial Intelligence} \\
			\textit{Universidad de Granada}\\
			Granada, Spain \\
			jopalacion@correo.ugr.es
		}
		\and
		\IEEEauthorblockN{Fernando Berzal}
		\IEEEauthorblockA{
			\textit{Dept. Computer Science and Artificial Intelligence} \\
			\textit{Universidad de Granada}\\
			Granada, Spain \\
			fberzal@acm.org
		}
	}
	\maketitle
	
	\begin{abstract}
	Community detection is a fundamental problem in the analysis of complex networks. It is the analogue of clustering in network data mining. Within community detection methods, hierarchical algorithms are popular. However, their iterative nature and the need to recompute the structural properties used to split the network (i.e. edge betweenness in Girvan and Newman's algorithm), make them unsuitable for large network data sets. In this paper, we study how local structural network properties can be used as proxies to improve the efficiency of hierarchical community detection while, at the same time, achieving competitive results in terms of modularity. In particular, we study the potential use of the structural properties commonly used to perform local link prediction, a supervised learning problem where community structure is relevant, as nodes are prone to establish new links with other nodes within their communities. In addition, we check the performance impact of network pruning heuristics as an ancillary tactic to make hierarchical community detection more efficient.\\ 
	\end{abstract}
	

	\section{Introduction}
	
	Networks are pervasive and the analysis of networks, therefore, is crucial in many scientific endeavors, from molecular biology and sociology to epidemiology and, why not, computer science. Community detection, defined as an unsupervised machine learning task, is a key problem within network data mining. Many community detection techniques have been proposed to detect communities in complex networks \cite{fortunato_community_2010}, but efficiency was, and still is, a great challenge for community detection algorithms \cite{liu_fast_2016}.
	\\\\
	Community detection methods can be classified attending to the kind of communities they can detect (overlapping or not), the criteria they use to detect their existence (local or global), the measure they try to optimize either explicitly or implicitly (e.g. modularity), or the algorithmic technique they resort to (hierarchical, partitioning, optimization-based, or spectral, just to name a few).
	\\\\
	Hierarchical community detection, as its traditional counterpart, hierarchical clustering, is one of the most popular techniques. Newman and Girvan's algorithm \cite{girvan_community_2002} is a divisive hierarchical algorithm for community detection that relies on edge betweenness to identify links that serve as bridges for different communities. Unfortunately, every time such a link is broken, edge betweenness has to be recomputed and this operation is costly. Radicchi \cite{radicchi_defining_2004} proposed a more efficient algorithm based on the clustering coefficient, which can be computed locally and leads to improved efficiency.
	\\\\
	In parallel to the development of unsupervised community detection algorithms, researchers also worked on supervised problems such as link prediction \cite{martinez_survey_2016}. As its name implies, link prediction methods try to predict the existence (or future formation) of a link between a pair of nodes. Many techniques have been proposed to solve this problem. Some of them resort to network structural properties, either local or global. For obvious reasons, local properties lead to more efficient link prediction algorithms, as happened with community detection methods. 
	\\\\
	In this paper, we analyze how such local structural properties, originally devised for a supervised learning problem, can be employed within an unsupervised learning technique to achieve competitive results and lead to more efficient hierarchical community detection algorithms.
	\\\\
	Our paper is organized as follows. In Section 2, we introduce hierarchical methods for community detection. In Section 3, we describe the link prediction problem, focusing on local link prediction, and survey some of the network structural properties that have been found to be useful for link prediction. In Section 4, we present the results of the experiments we have performed on some benchmark network data sets. Finally, Section 5 ends our paper with some conclusions and recommendations on the use of local structural properties for community detection.

	\section{Hierarchical community detection}
	
	Hierarchical clustering, a traditional unsupervised machine learning technique, proceeds iteratively until it creates a complete hierarchy of clusters \cite{liu_web_2011}. The resulting hierarchy is typically displayed by a tree, called dendrogram, where leaves represent individual instances and the root of the tree covers the whole data set. 
	\\\\
	The clustering process can follow either a bottom-up approach, from the leaves to the root, or a top-down approach, from the root to the leaves. In the former case, the clustering algorithm is agglomerative. In the latter case, it is divisive. Since we usually start with a large data set and we are typically interested in a small number of clusters (i.e. communities when we are dealing with networks), divisive algorithms are often preferred.

	\subsection{Girvan and Newman's community detection method}
	
	Girvan and Newman \cite{girvan_community_2002} proposed a divisive hierarchical clustering algorithm for networks. Their algorithm is based on the idea of edge betweenness as a suitable measure for detecting links that connect different communities. Edge betweenness counts the number of shortest paths that traverse each edge (or link). The larger that number, the higher the probability that the link connects different communities. Hence, edge betweenness can be used to suggest the order in which links can be cut in a conventional divisive hierarchical clustering algorithm.
	\\\\
	Formally, edge betweenness is defined for each edge $e\in E$ as follows \cite{newman_finding_2004}:
	
	\begin{equation*}
	C_B(e)=\sum_{u,v\in{V}}\dfrac{g_e(u,v)}{g(u,v)}\label{eq1}
	\end{equation*}\	
	Where $g(u,v)$ is the total number of minimum paths between nodes $u$ and $v$, whereas $g_e (u,v)$ is the number of minimum paths between nodes $u$ and $v$ that pass through node $e$. 
	\\\\
	Small edge betweenness values indicate that edges probably belong to the same community. In contrast, those edges that connect nodes from different communities will have higher edge betweenness values \cite{newman_finding_2004}.
	\\\\
	Girvan and Newman's algorithm starts by computing the edge betweenness for all the edges in the network. Then, iteratively, while there exists an edge whose betweenness is above an user-provided threshold, the edge with the highest betweenness is removed from the network and edge betweenness in recalculated throughout the network.
	\\
	\begin{figure}[htbp]
		\centerline
		{
			\includegraphics[scale=1]{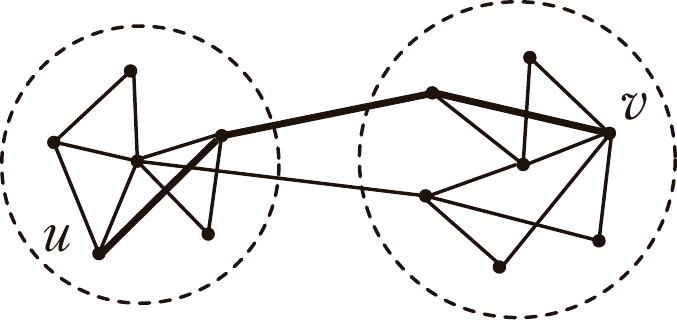}
		}
		\caption{Identification of edges with high edge betweenness \cite{newman_networks:_2010}.}
		\label{figura1}
	\end{figure}
	
	One of the major drawbacks of Girvan and Newman's algorithm is its computational complexity, since every iteration of the hierarchical algorithms requires recomputing the edge betweenness for all the edges in the network. In general, the algorithmic complexity of the algorithm is $O(mn(m+n))$ for a network with $m$ edges and $n$ nodes. In other words, it can reach $O(n^3)$ for dense networks, where $m\propto n$. Therefore, this method is not recommended for large networks \cite{newman_networks:_2010} despite its good results and intuitive appeal.

	\subsection{Radicchi's community detection method}
	
	Given the computational complexity of Girvan and Newman's algorithm, Radicchi\cite{radicchi_defining_2004} proposed a more efficient algorithm based on a simple idea:
	\\\\
	A community contains heavily-connected nodes. Therefore, cycles are frequent within a community. However, links that connect different communities are involved in a lower number of cycles. In other words, edges connecting different communities are less likely to be involved in cycles \cite{newman_detecting_2004}.
	\\\\
	Following this observation, Radicchi's algorithm is based on the definition of an edge clustering coefficient. This coefficient is analogous to the usual node clustering coefficient, which is the number of triangles including a node divided by the number of possible triangles that could potentially be formed with its neighbors. The edge clustering coefficient is, therefore, the number of triangles to which a given edge belongs, divided by the number of triangles that might potentially include it, given the degrees of its adjacent nodes.
	
	\begin{figure}[htbp]
		\centerline
		{
			\includegraphics[scale=2.0]{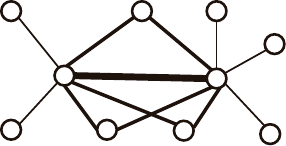}
		}
		\caption{Schematic illustration of the edge clustering coefficient introduced by Radicchi \cite{fortunato_community_2010}.}
		\label{figura20}
	\end{figure}

	Formally, the edge clustering coefficient for the edge connecting node $i$ to node $j$ is defined as \cite{newman_detecting_2004}:
	
	\begin{equation*}
	C_{ij}=\dfrac{z_{ij}+1}{\min(k_i-1,k_j-1)}\label{eq100}
	\end{equation*}\\
	where $z_{ij}$ is the number of triangles built on that edge (i.e. triangles containing the edge from node $i$ to node $j$) and $k_i$ is the degree of node $i$. A problem arises when the number of triangles is zero ($z_{ij}=0$ when $\min(k_i-1,k_j-1)=0$), and the degenerate case is removed by adding one to the number of triangles in the numerator.
	\\\\
	Instead of considering all possible cycles, Radicchi focuses on the shortest ones, triangles of length 3. Edges that connect different communities tend to exhibit a small edge clustering coefficient (i.e. they take part in in few or no triangles at all). Hence, the edge clustering coefficient is a good measure for detecting bridges between communities.
	\\\\
	Whereas Girvan and Newman's algorithm chose edges with the highest edge betweenness, Radicchi's algorithm selects those with the lowest edge clustering coefficient. 
	\\\\
	Girvan and Newman's algorithm was computationally expensive because it required the repeated computation of a global network property, the edge betweenness for each edge in the network (a property whose value depends on the properties of the whole network, which change every time we remove and edge). However, Radicchi's algorithm is based on a local property, the edge clustering coefficient, and is much faster.
	\\\\	
	The algorithmic complexity of Radicchi's hierarchical community detection algorithms is $O(n^2)$. However, the algorithm tends to fail if the average clustering coefficient of the network is small and, then, the edge clustering coefficient is small for all edges \cite{newman_networks:_2010}. 
	\\\\	
	Radicchi's algorithm works well in networks with many triangles, e.g. social networks. However, in technological or biological networks, there are fewer of these structures, which makes it difficult to identify the edges that bridge different communities \cite{newman_networks:_2010}.

	\section{Local link prediction}
	
	Link prediction, a widely-studied supervised learning problem, can be useful for identifying potential candidates to improve the efficiency of Girvan and Newman's algorithm and overcome the limitations of Radicchi's method. Link prediction explores the intrinsic features of nodes and edges and their behavior in the past or in the future \cite{clauset_hierarchical_2008}.
	\\\\	
	Many techniques have been proposed for the prediction of links in complex networks: local, global, quasi-local, classification-based, metaheuristic, and factor-based methods \cite{martinez_survey_2016}.
	\\\\	
	Some studies by Liben-Nowell and Kleinberg \cite{liben-nowell_link_2003} and Zhou et al. \cite{zhou_predicting_2009} confirm that local link prediction methods offer good results with a limited computational complexity. Their results might not be as accurate as those obtained by global methods, but global methods are computationally costly \cite{valverde-rebaza_link_2012}.	
	\\\\	
	Obviously, local link prediction methods will never be able to predict the existence of links between remote nodes. However, most links appear within a given neighborhood and, given the computational advantage they offer, local link prediction techniques are ideal candidates for improving the performance of existing hierarchical community detection methods.
	\\\\	
	In the following paragraphs, we briefly survey some of the criteria used by local link prediction methods. They often define a similarity measure $s(x,y)$  between two nodes, $x$ and $y$, to rank potential links and their computational complexity is usually $O(vk^3)$, where $v$ is the number of nodes and $k$ is the maximum degree of the nodes \cite{martinez_survey_2016}. 
	
	\subsection{Common neighbors (CN)}
	
	The simplest local link prediction method counts the number of neighbors in common between a pair of nodes $(x,y)$. The greater the number of common neighbors, the greater the probability of formation of a link between them \cite{lu_link_2011}:
	\begin{equation*}
	s(x,y)=|\Gamma_x \cap \Gamma_y|
	\end{equation*}
	where, for a node $x$, $\Gamma_x$ indicates the set of neighbors of $x$.
	
	
	\subsection{The Adamic-Adar index (AA)}
	
	The Adamic-Adar index evaluates the level of binding between a pair of nodes from the number of neighbors they share, yet the contribution of each neighbor is penalized by its degree \cite{adamic_friends_2003}:
	\begin{equation*}
	s(x,y)=\sum_{z\in\Gamma_x\cap\Gamma_y}\dfrac{1}{\log|\Gamma_z|}
	\end{equation*}
	
	\subsection{The resource allocation index (RA)}
	
	The resource allocation index is motivated by the resource allocation process that takes place in complex networks \cite{zhou_predicting_2009}. It models the transmission of units of resources between two unconnected nodes $x$ and $y$ through their neighbors: each neighborhood node gets a unit of resource from $x$ and equally
	distributes it to its neighbors. The amount of resources obtained by node $y$ is, therefore, given by: 
	\begin{equation*}
	s(x,y)=\sum_{z\in\Gamma_x\cap\Gamma_y}\dfrac{1}{|\Gamma_z|}
	\end{equation*}
	
	\subsection{The preferential attachment index (PA)}
	
	The preferential attachment index is a direct result of the well-known Barab\'asi-Albert complex network formation model \cite{barabasi_emergence_1999}. In that model, which leads to a power law degree distribution, the probability of link formation between two nodes increases as the degree of these nodes does:
	\begin{equation*}
	s(x,y)=|\Gamma_x||\Gamma_y|
	\end{equation*}\
	
	\subsection{The Jaccard index (JA)}
	
	The Jaccard index was proposed to compare the similarity and diversity of two sets \cite{jaccard_etude_1901}. In our case, the sets of neighbors:
	\begin{equation*}
	s(x,y)=\dfrac{|\Gamma_x\cap\Gamma_y|}{|\Gamma_x\cup\Gamma_y|}
	\end{equation*}\
	
	\subsection{The Sørensen index (SO)}
	
	The Sørensen index also assesses the similarity between two sets \cite{sorensen_method_1948}. It is very similar to the Jaccard index, yet less sensitive to outliers:
	\begin{equation*}
	s(x,y)=\dfrac{2|\Gamma_x\cap\Gamma_y|}{|\Gamma_x|+|\Gamma_y|}
	\end{equation*}\ 
	
	\subsection{The Salton index (SA)}
	
	The Salton index \cite{salton_introduction_1986}, also known as cosine similarity, is closely-related to the Jaccard index:
	\begin{equation*}
	s(x,y)=\dfrac{|\Gamma_x\cap\Gamma_y|}{\sqrt{|\Gamma_x|+|\Gamma_y|}}
	\end{equation*}\
	
	\subsection{The hub promoted index (HP)}
	
	The hub promoted index was the result of a study of modularity in metabolic networks \cite{ravasz_hierarchical_2002}. Such networks show a hierarchical structure with small highly internally connected modules that are also highly isolated from each other. The HP index avoids link formation between hub nodes and promotes link formation between low-degree nodes and hubs:
	\begin{equation*}
	s(x,y)=\dfrac{|\Gamma_x\cap\Gamma_y|}{\min(|\Gamma_x|,|\Gamma_y|)}
	\end{equation*}\
	
	\subsection{The hub depressed index (HD)}
	
	The hub depressed index was also a result of the same study \cite{ravasz_hierarchical_2002}. The HD index promotes link formation between hubs and between low-degree nodes, but not between hubs and low-degree nodes:
	\begin{equation*}
	s(x,y)=\dfrac{|\Gamma_x\cap\Gamma_y|}{\max(|\Gamma_x|,|\Gamma_y|)}
	\end{equation*}\
	
	\subsection{The local Leicht-Holme-Newman index (LLHN)}
	
	The LLHN index is defined as the ratio of actual paths of length two between two nodes and a value proportional to the expected number of paths of length two between them. It is similar to the Jaccard and Salton indices, yet more sensitive as a measure of structural equivalence \cite{leicht_vertex_2006}:
	\begin{equation*}
	s(x,y)=\dfrac{|\Gamma_x\cap\Gamma_y|}{|\Gamma_x||\Gamma_y|}
	\end{equation*}\
	
	\section{Experimentation}
	
	In our experimentation, we will analyze the results obtained by a hierarchical community detection algorithm using a dozen different criteria on half a dozen benchmark data sets. As baseline, we will resort to the edge betweenness of Girvan and Newman's method as well as the edge clustering coefficient of Radicchi's proposal. We will also evaluate the performance of the 10 different criteria used by local link prediction methods described in the previous Section: common neighbors (CN), the Adamic-Adar index (AA), the resource allocation index (RA), the preferential attachment index (PA), the hub depressed index (HD), the hub promoted index (HP), the Jaccard index (JA), the local Leicht-Holme-Newman (LLHN) index, the Salton index (SA), and the Sørensen index (SO). \\
	
	\subsection{Benchmark data sets}
	
	The following network data sets were obtained from networkrepository.com \cite{rossi_network_2015}, SNAP \cite{leskovec_snap_2014}, and Pajek datasets \cite{batagelj_pajek_2006}:
	
	\begin{itemize}
		\item Zachary’s karate club \cite{zachary_information_1977}
		\item Word adjacencies \cite{newman_finding_2006}
		\item Dolphin \cite{lusseau_bottlenose_2003}
		\item Les Misérables \cite{knuth_stanford_1993}
		\item U.S. politics \cite{eakin_study_2004}\cite{krebs_political_2008}
		\item American football \cite{girvan_community_2002}\\
	\end{itemize}
	
	Some key structural properties of those networks are shown in Table \ref{tabla1}.
	\subsection{Data}
	
	\begin{table}[htbp]
		\caption{Key structural properties of the networks used in our experiments: number of nodes ($n$), number of links/edges ($m$), average node  degree ($\langle k \rangle$), and average clustering coefficient ($C$).}
		\begin{center}
			\scalebox{0.9}{
				\begin{tabular}{lrrrrc}
					\hline
					Network & $n$ & $m$ & $\langle k \rangle$ & $C$ & reference\\
					\hline
					Zachary’s karate club & 34 & 78 & 4.588 & 0.426 & \cite{zachary_information_1977}\\
					Word adjacencies & 112 & 425 & 7.528 & 0.403 & \cite{newman_finding_2006}\\
					Dolphin & 62 & 159 & 5.129 & 0.307 & \cite{lusseau_bottlenose_2003}\\
					Les Misérables & 77 & 254 & 6.597 & 0.389 & \cite{knuth_stanford_1993}\\
					US politics & 105 & 441 & 8.400 & 0.330 & \cite{eakin_study_2004}\cite{krebs_political_2008}\\
					American Football & 115 & 613 & 10.661 & 0.399 & \cite{girvan_community_2002}\\
					\hline				
				\end{tabular}
			}
		\end{center}
		\label{tabla1}
	\end{table}

	\subsection{Edge selection criteria}
	
	The dozen different versions of the hierarchical community detection algorithm were implemented in Python with the help of the following libraries: NumPy \cite{walt_numpy_2011} for mathematical computations, NetworkX \cite{hagberg_exploring_2008} and NOESIS \cite{martinez_noesis_2015}\cite{martinez_noesis_2019} for network data mining algorithms.
	\\
	\begin{table*}[htbp]
		\centering
		\caption{Lineal correlation between edge betweenness and the different criteria used by local link prediction techniques.}
		\resizebox{16cm}{!} {
			\begin{tabular}{p{1.5cm}lllllllllll}
				\hline
				\textbf{Network}      & \textbf{CN}    & \textbf{AA}    & \textbf{RA}    & \textbf{PA}    & \textbf{HD}    & \textbf{HP}    & \textbf{JA}    & \textbf{LLHN}  & \textbf{SA}    & \textbf{SO}    & \textbf{Avg.}  \\ \hline
				Zachary 
				&-0.201          &-0.107          &-0.059          & 0.392          &-0.507          &-0.326          &-0.481          &-0.580          &-0.477          &-0.492          & \textbf{-0.284} \\ 
				Word                  
				& 0.068          & 0.103          & 0.134          & 0.239          &-0.330          & 0.017          &-0.254          &-0.384          &-0.196          & 0.266          & \textbf{-0.087} \\ 
				Dolphin               
				&-0.431          &-0.410          &-0.384          &-0.098          &-0.469          &-0.490          &-0.465          &-0.466          &-0.486          &-0.484          & \textbf{-0.418} \\ 
				Misérables        
				&-0.338          &-0.280          &-0.204          & 0.380          &-0.547          &-0.497          &-0.539          &-0.508          &-0.561          &-0.559          & \textbf{-0.365} \\ 
				Politics
				&-0.271          &-0.240          &-0.196          &-0.010          &-0.397          &-0.424          &-0.418          &-0.343          &-0.455          &-0.437          & \textbf{-0.319} \\ 
				Football 
				&-0.777          &-0.779          &-0.780          &-0.016          &-0.785          &-0.787          &-0.757          &-0.790          &-0.787          & 0.787          & \textbf{-0.704} \\ 
				\textbf{Average}      & \textbf{-0.325} & \textbf{-0.286} & \textbf{-0.248} & \textbf{0.148} & \textbf{-0.506} & \textbf{-0.418} & \textbf{-0.486} & \textbf{-0.512} & \textbf{-0.494} & \textbf{-0.504} & \textbf{}      \\ \hline
			\end{tabular}
		}
		\label{tabla2}
	\end{table*}
	
	First of all, we measured the correlation between edge betweenness (the criterion used by the original Girvan and Newman's algorithm) and the 10 different local criteria used by local link prediction techniques. The correlation is often negative, given that we choose edges that maximize edge betweenness, whereas we will choose edges that minimize the local structural property defined by the corresponding local link prediction method. However, correlation is usually weak, only stronger for the American Football network.
	\\
	\subsection{Community detection results}
	
\begin{figure}[htbp]
		\centering
		\subfigure[adjnoun]{\includegraphics[width=42mm]{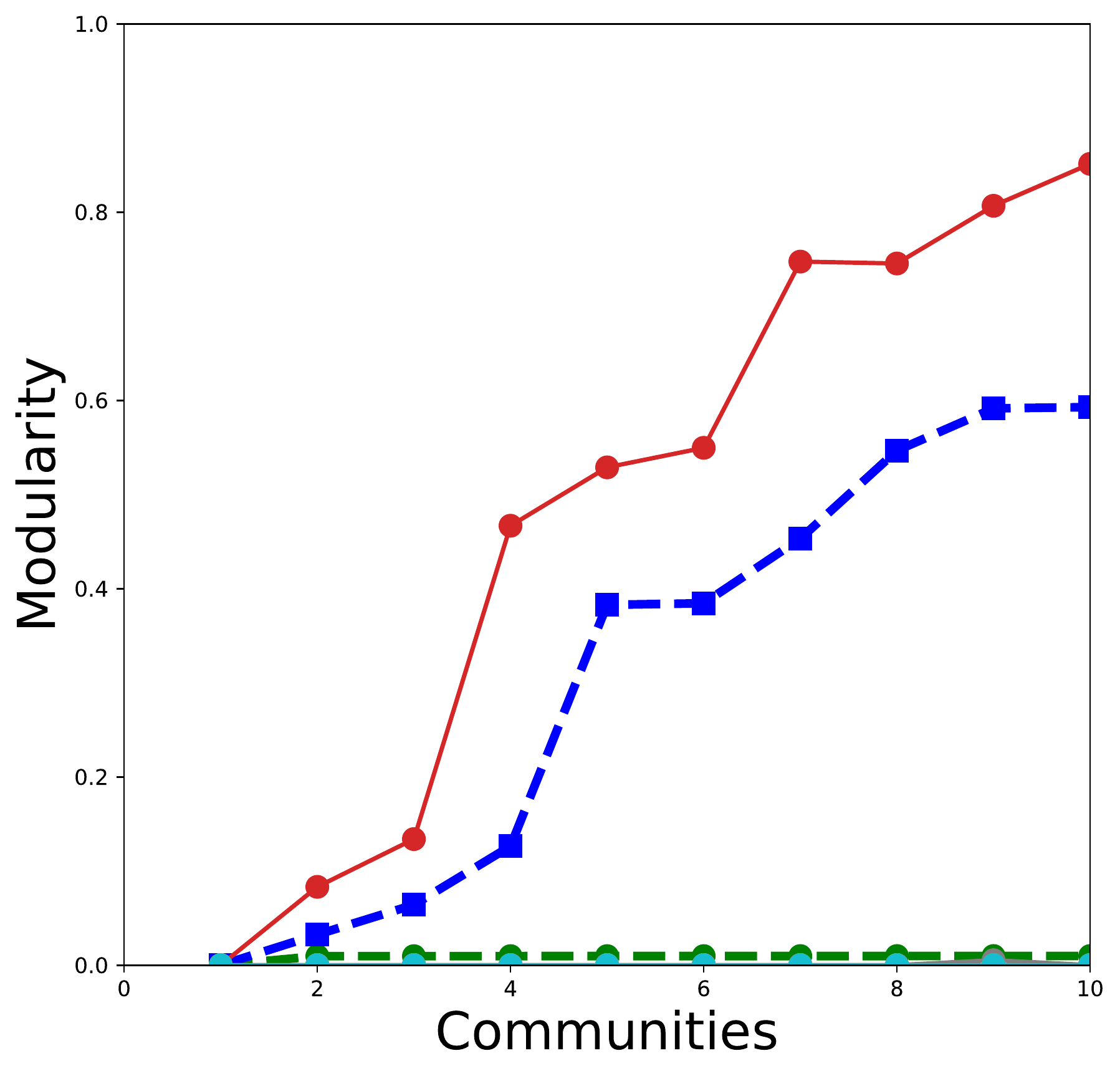}}\hspace{2mm}
		\subfigure[dolphins]{\includegraphics[width=42mm]{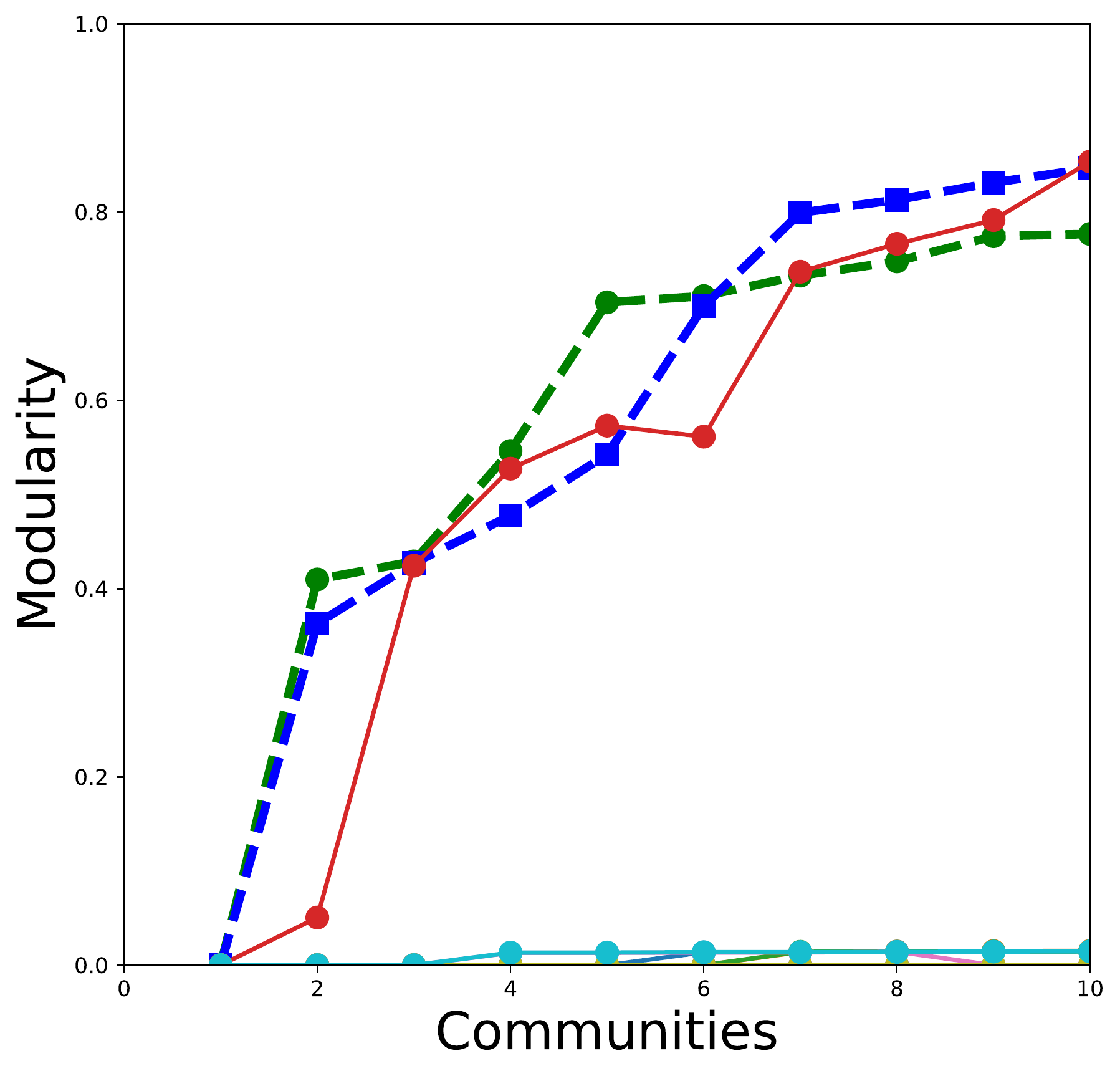}}\vspace{2mm}
		\subfigure[football]{\includegraphics[width=42mm]{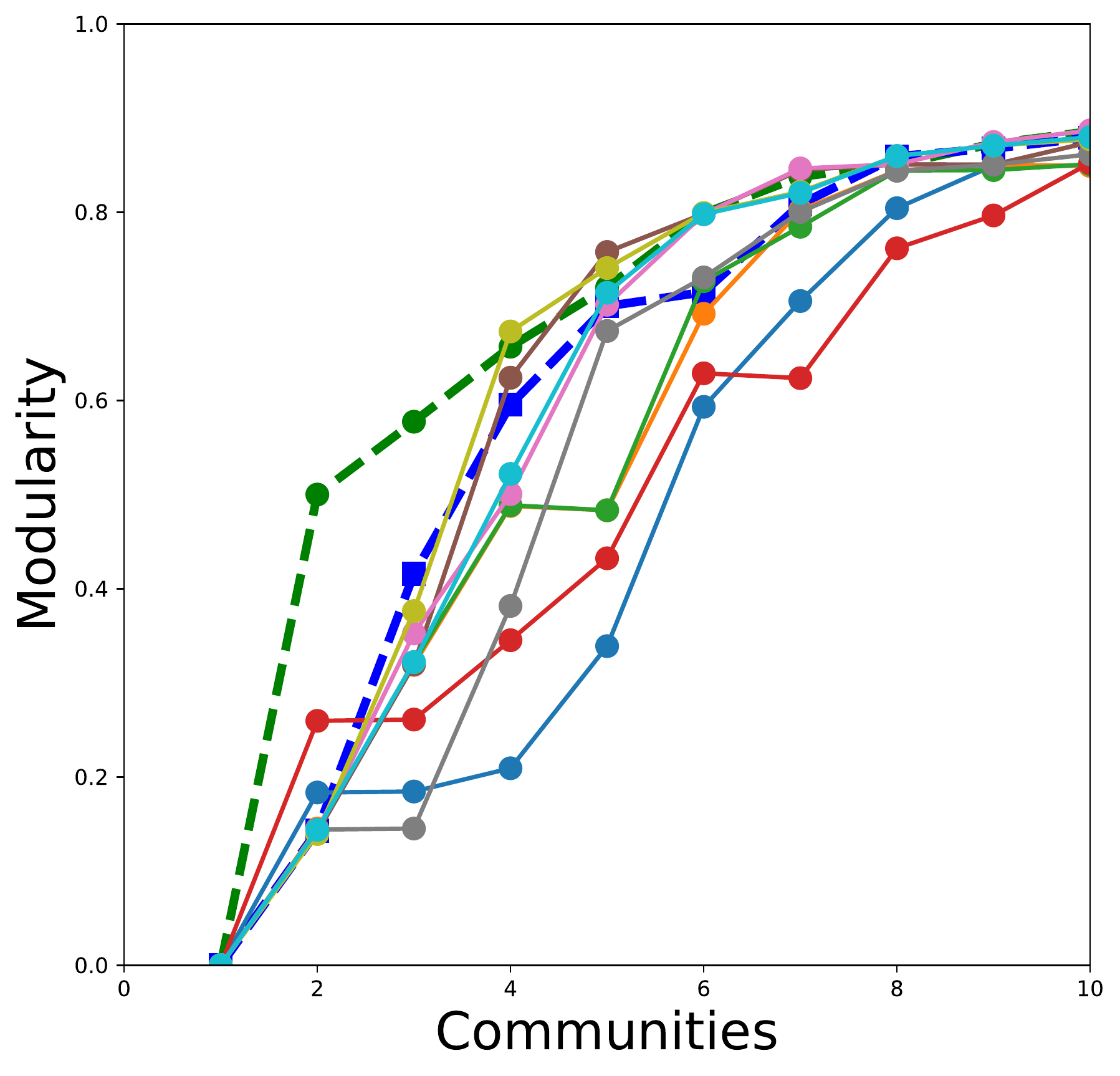}}\hspace{2mm}
		\subfigure[karate]{\includegraphics[width=42mm]{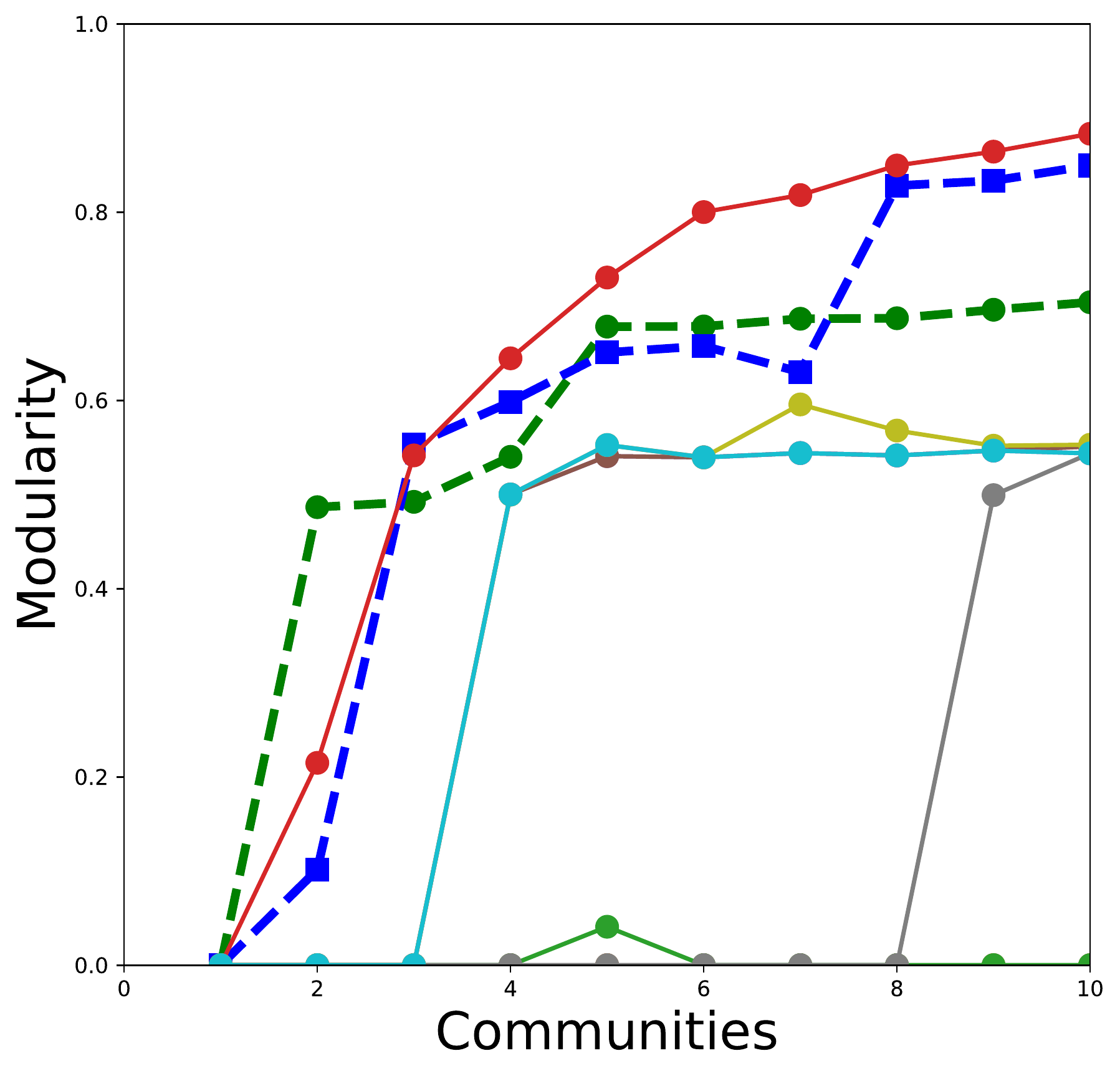}}\vspace{2mm}		\subfigure[lesmis]{\includegraphics[width=42mm]{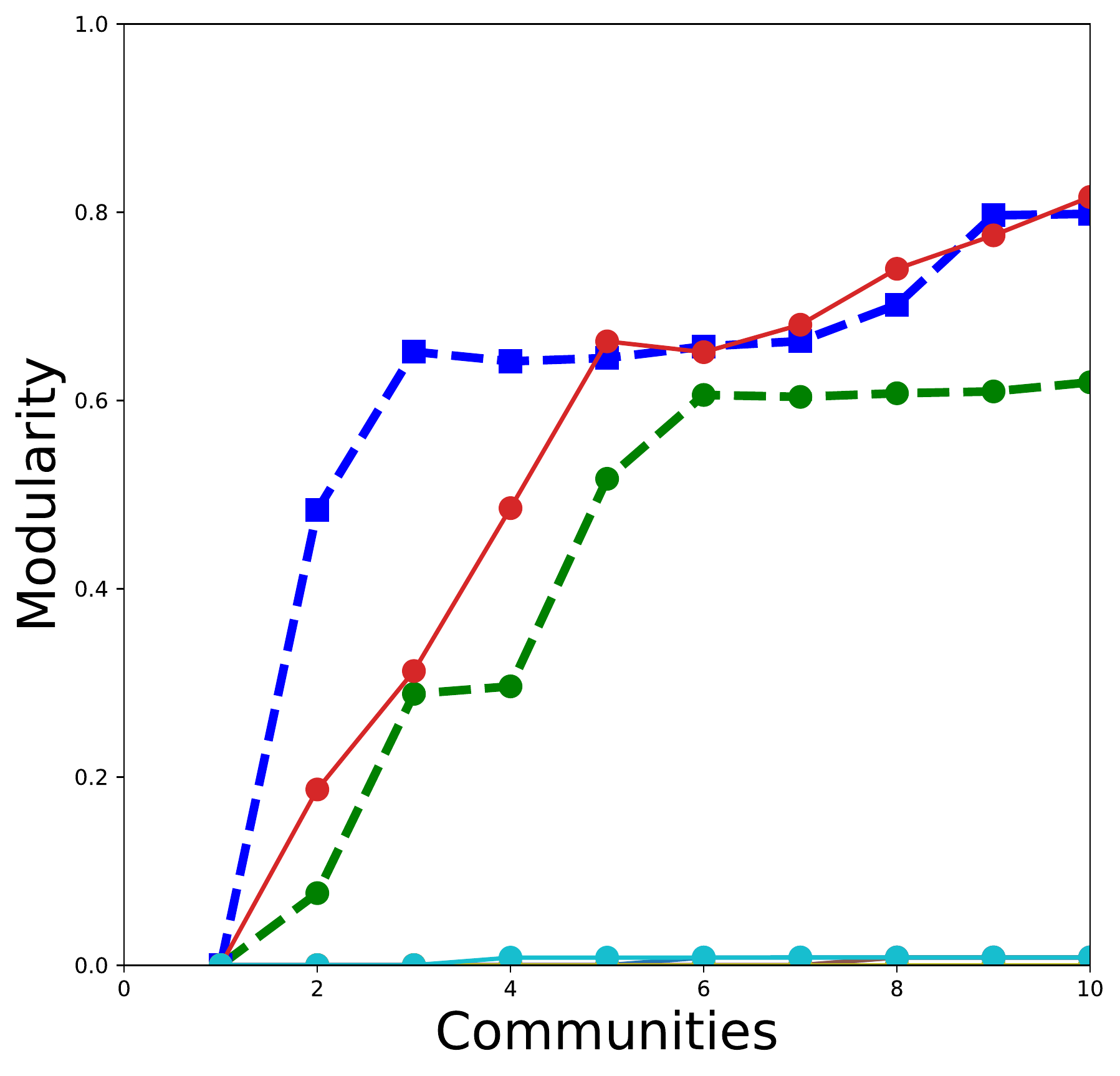}}\hspace{2mm}
		\subfigure[polbooks]{\includegraphics[width=42mm]{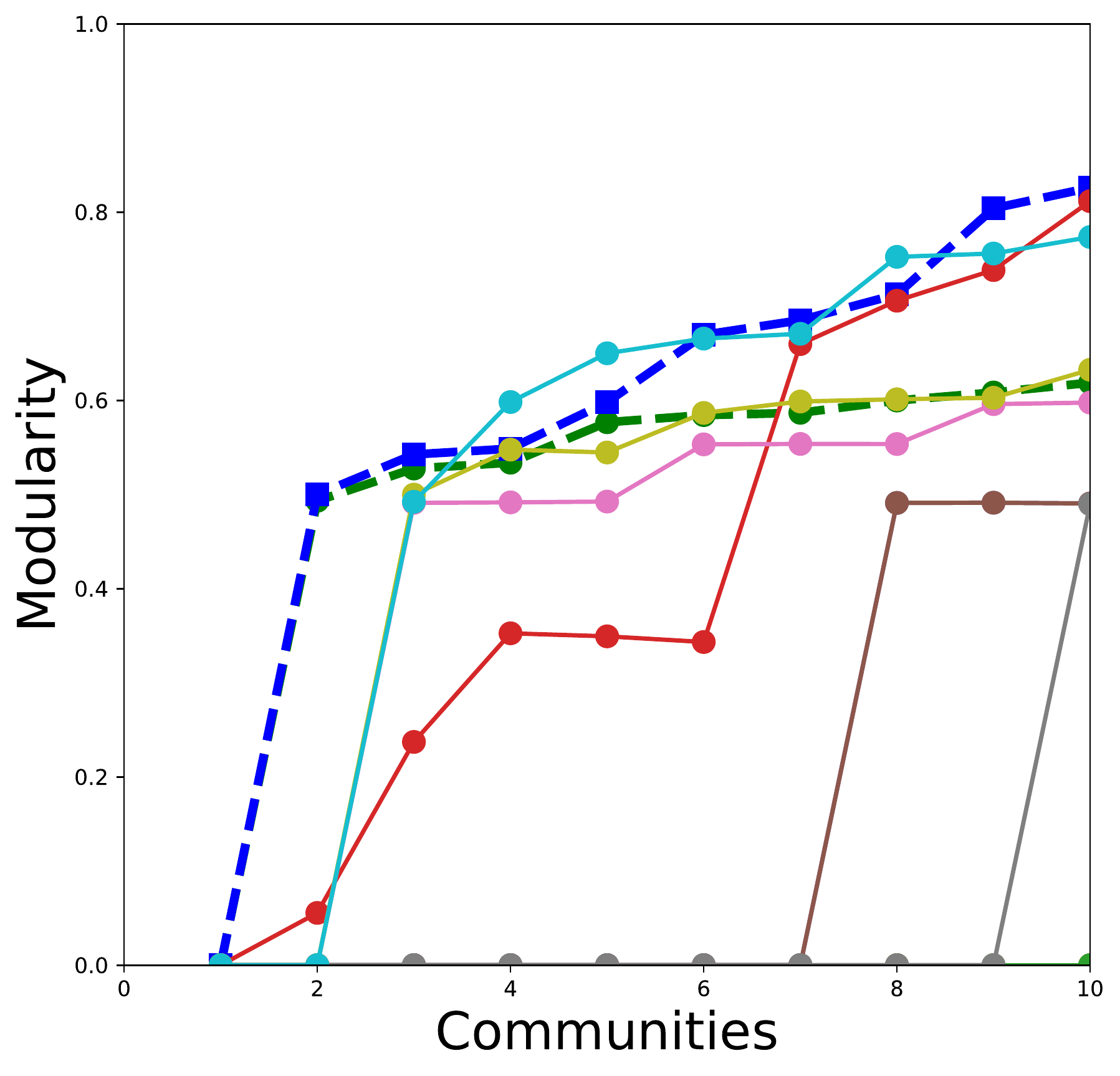}}\vspace{2mm}
		\subfigure{\includegraphics[width=90mm]{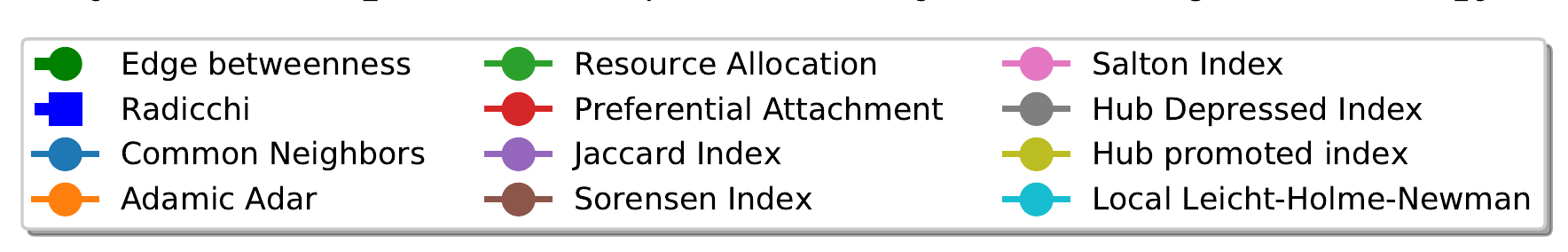}}\vspace{2mm}
		\caption{Modularity results using 12 different criteria for hierarchical community detection.} 
		\label{figura2}
	\end{figure}

	We used 12 different criteria in the hierarchical community detection algorithm and measured the results in terms of modularity. In optimization-based community detection algorithms, modularity makes reference to any numerical measure that is suitable for detecting communities in networks (i.e. the target of community detection as a numerical optimization problem). Here, however, we refer to the original modularity $Q$, as defined by Mark E.J. Newman \cite{newman_2006_modularity}.
	\\\\
	The less efficient Girvan and Newman's algorithm, which resorts to edge betweenness, is included as a baseline. Radicchi's edge-clustering-coefficient-based algorithm is also included in our comparison with 10 different local link prediction indicators.
	\\\\
	The results are displayed in Figure \ref{figura2}. The hierarchical clustering results from 2 to 10 communities are shown for 6 different networks in terms of Newman's $Q$ modularity. Results vary depending on the network properties. 
	\\\\
	In some networks, like adjnoun, only Radicchi's clustering coefficient and the preferential attachment index seem to work well (maybe due to the ``rich get richer'' effect observed in power law degree distributions that result from a preferential attachment model of network formation). In other networks, like football, where there is a strong community structure (regional divisions within a sport season), any measure is able to detect communities. In general, however, results vary and some measures fail to detect communities (even Girvan and Newman's edge betweenness).
	\\

	\subsection{Network pruning}
	
\begin{figure}[thbp]
		\centering
		\subfigure[adjnoun]{\includegraphics[width=42mm]{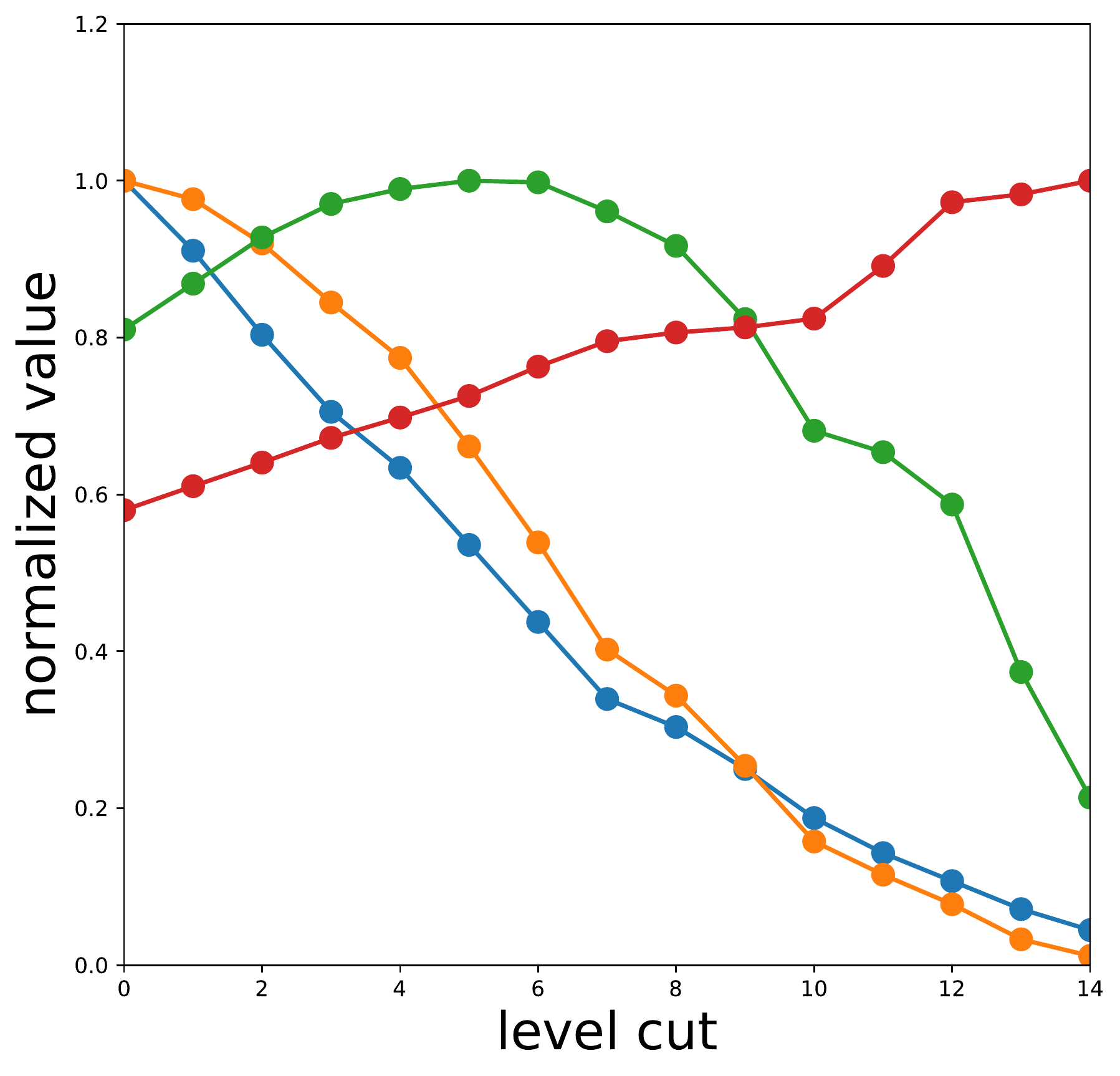}}\hspace{2mm}
		\subfigure[dolphins]{\includegraphics[width=42mm]{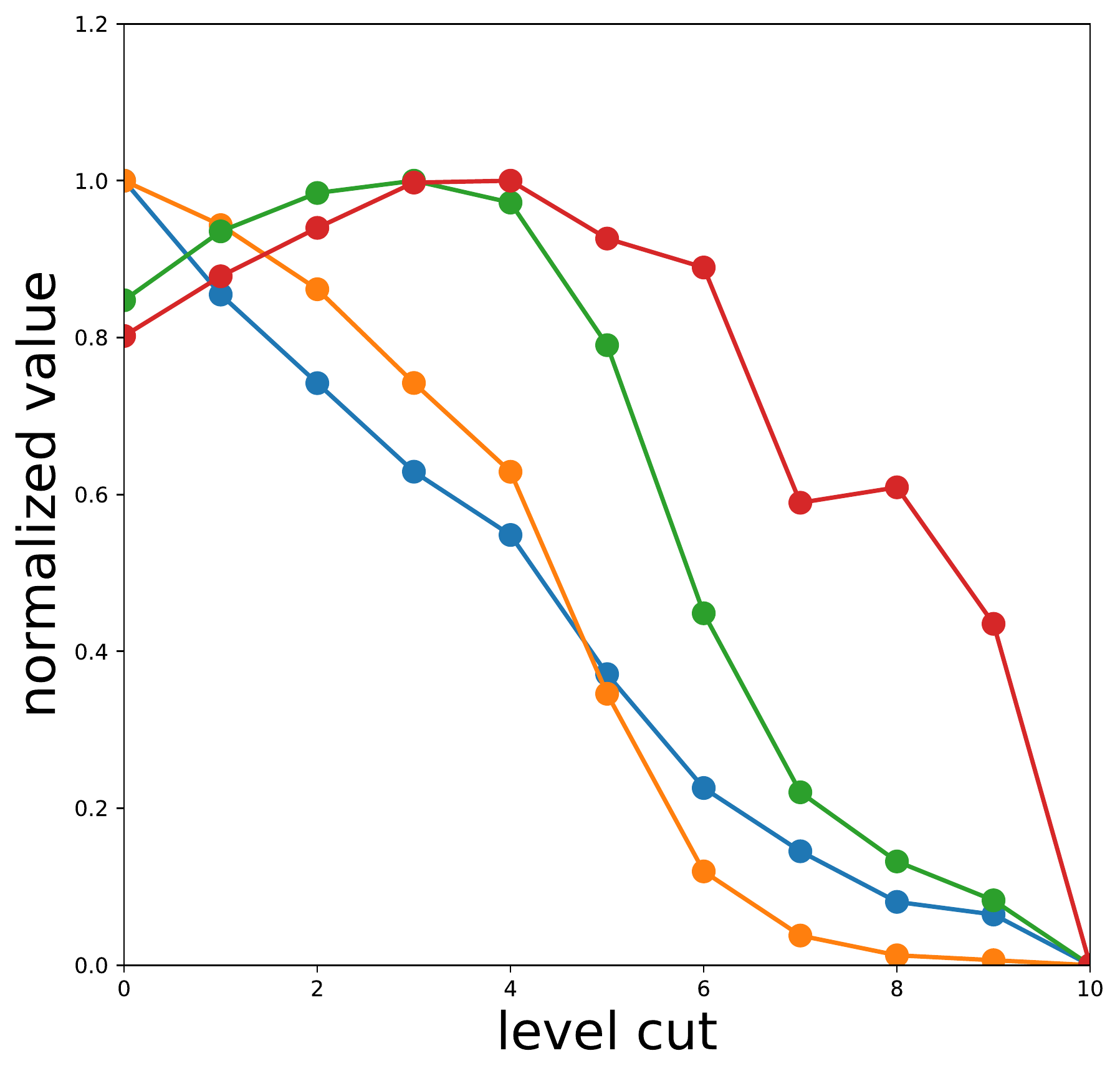}}\vspace{2mm}
		\subfigure[football]{\includegraphics[width=42mm]{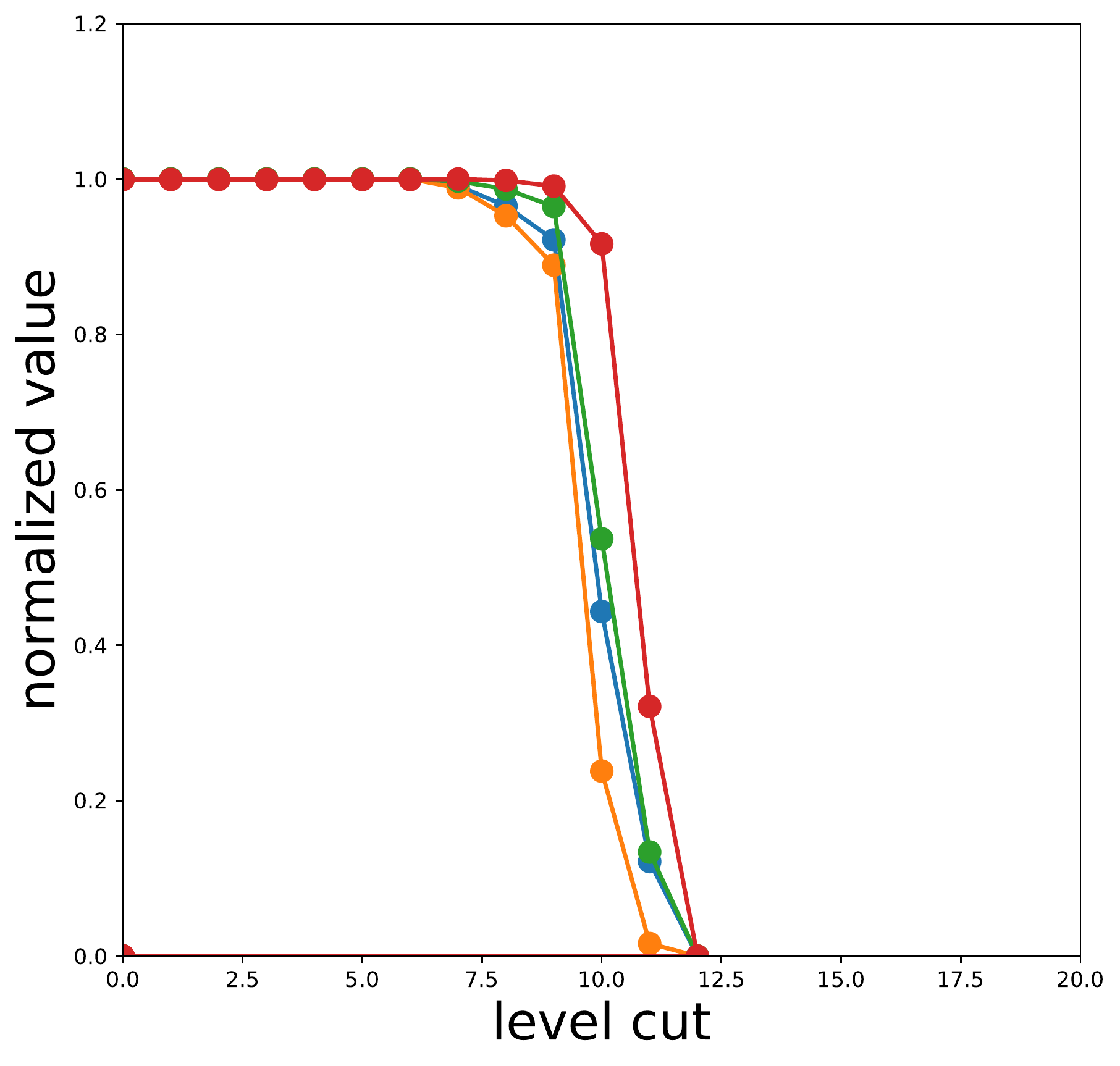}}\hspace{2mm}
		\subfigure[karate]{\includegraphics[width=42mm]{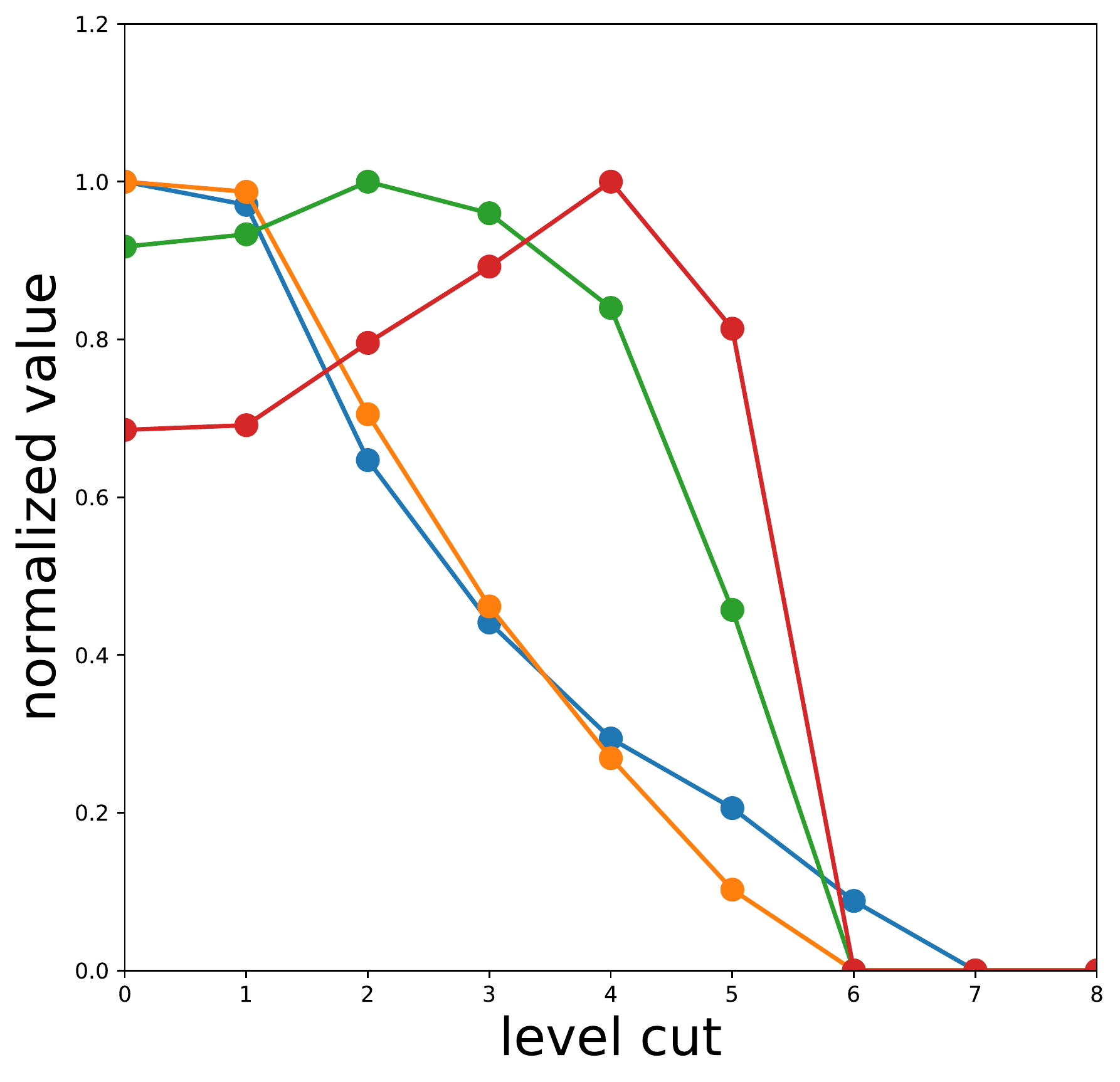}}\vspace{2mm}			\subfigure[lesmis]{\includegraphics[width=42mm]{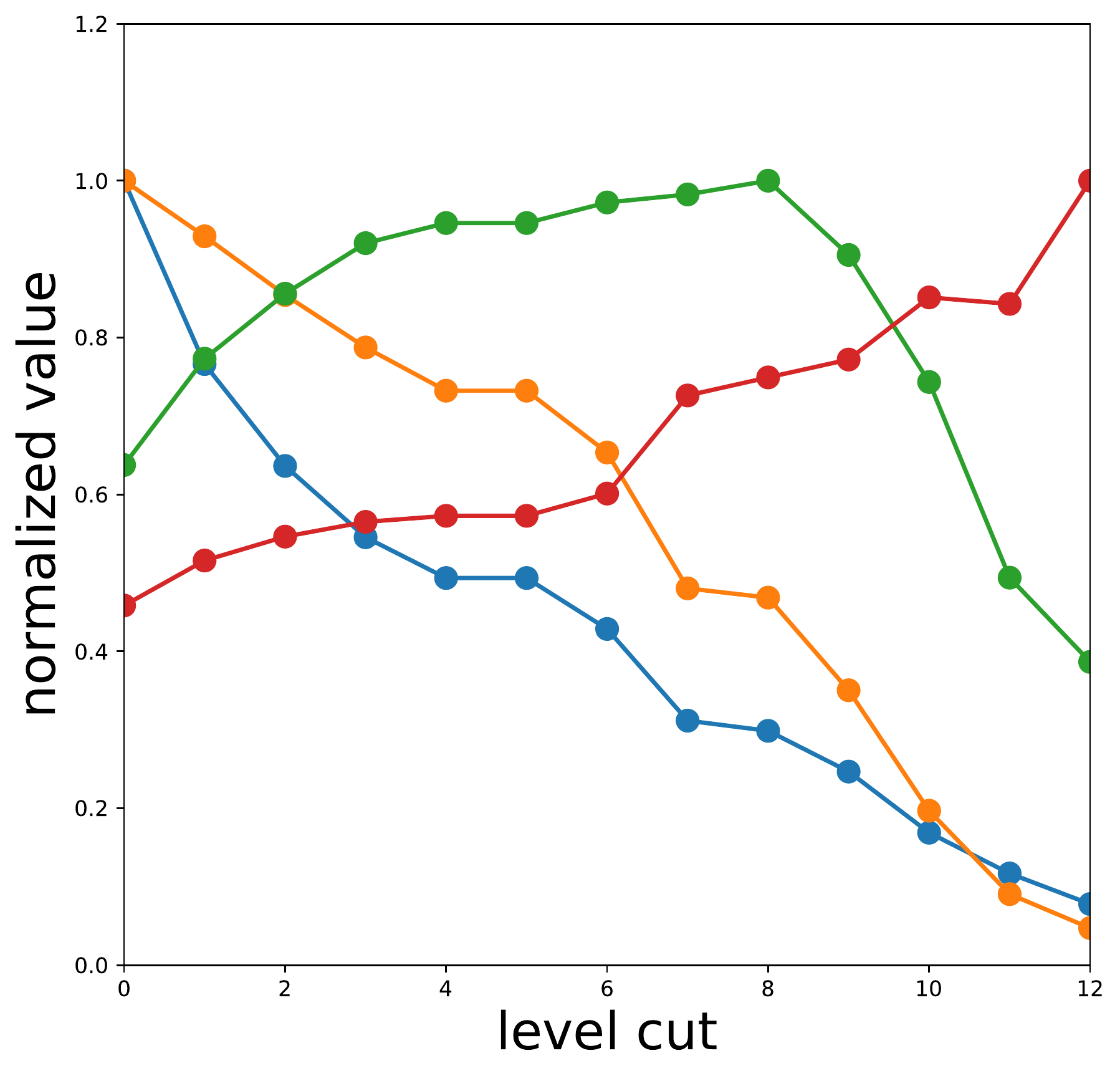}}\hspace{2mm}
		\subfigure[polbooks]{\includegraphics[width=42mm]{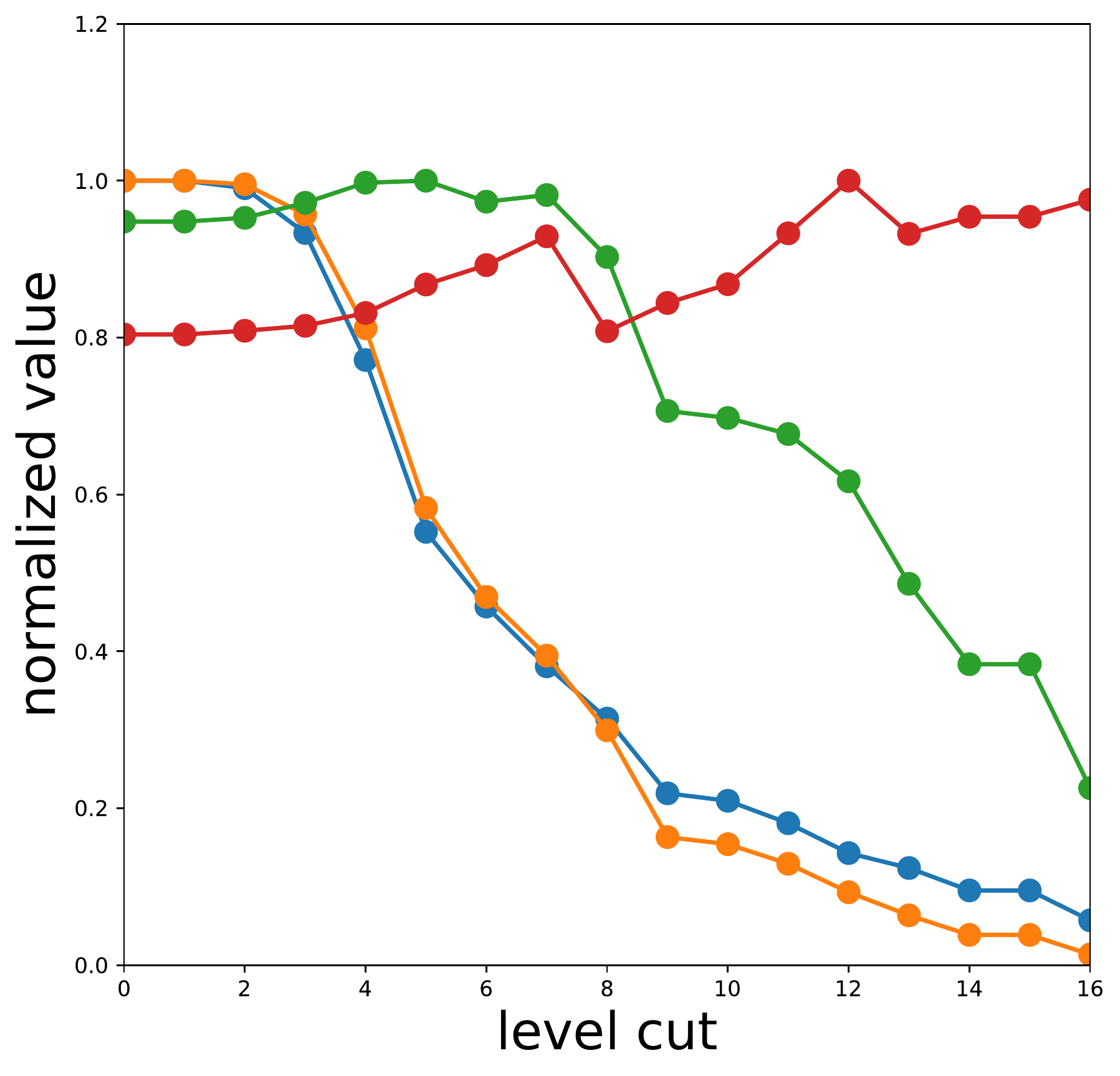}}\vspace{2mm}
		\subfigure{\includegraphics[width=90mm]{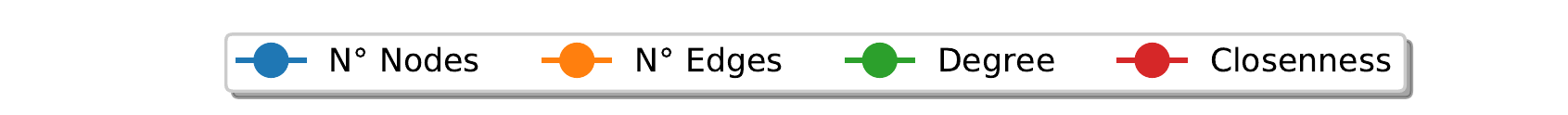}}\vspace{2mm}
		\caption{Evolution of network structural properties as more and more nodes are pruned from the original networks.} 
		\label{figura3}
	\end{figure}

	To improve the results discussed in the previous paragraphs, we proceeded to prune the networks in a controlled way. The motivation behind that pruning is that networks often include low degree nodes that do not truly participate in network community formation and those nodes might be adding noise to the measures of the local structural properties we use to split the networks.
	\\\\
	Our pruning algorithms consists of eliminating those nodes with a low degree, as well as their adjacent links. Obviously, we safeguard the data corresponding to those nodes and links, so that the community detection results are always reported on the whole network, not just on the pruned network. 
	\\\\
	We measured the impact of pruning at different levels, i.e. using different minimum degree thresholds, on the structural properties of networks. In Figure \ref{figura3}, we see how the average node degree and clustering coefficient evolve as more and more nodes and edges are removed from the original networks.
	\\
	
	The optimal pruning strategy must be determined individually, taking into account the specific features of each network. Excessive pruning could distort the topology of the network when carelessly performed.
	
	\subsection{Improved community detection results with network pruning}
	
\begin{figure}[htbp]
		\centering
		\subfigure[adjnoum]{\includegraphics[width=42mm]{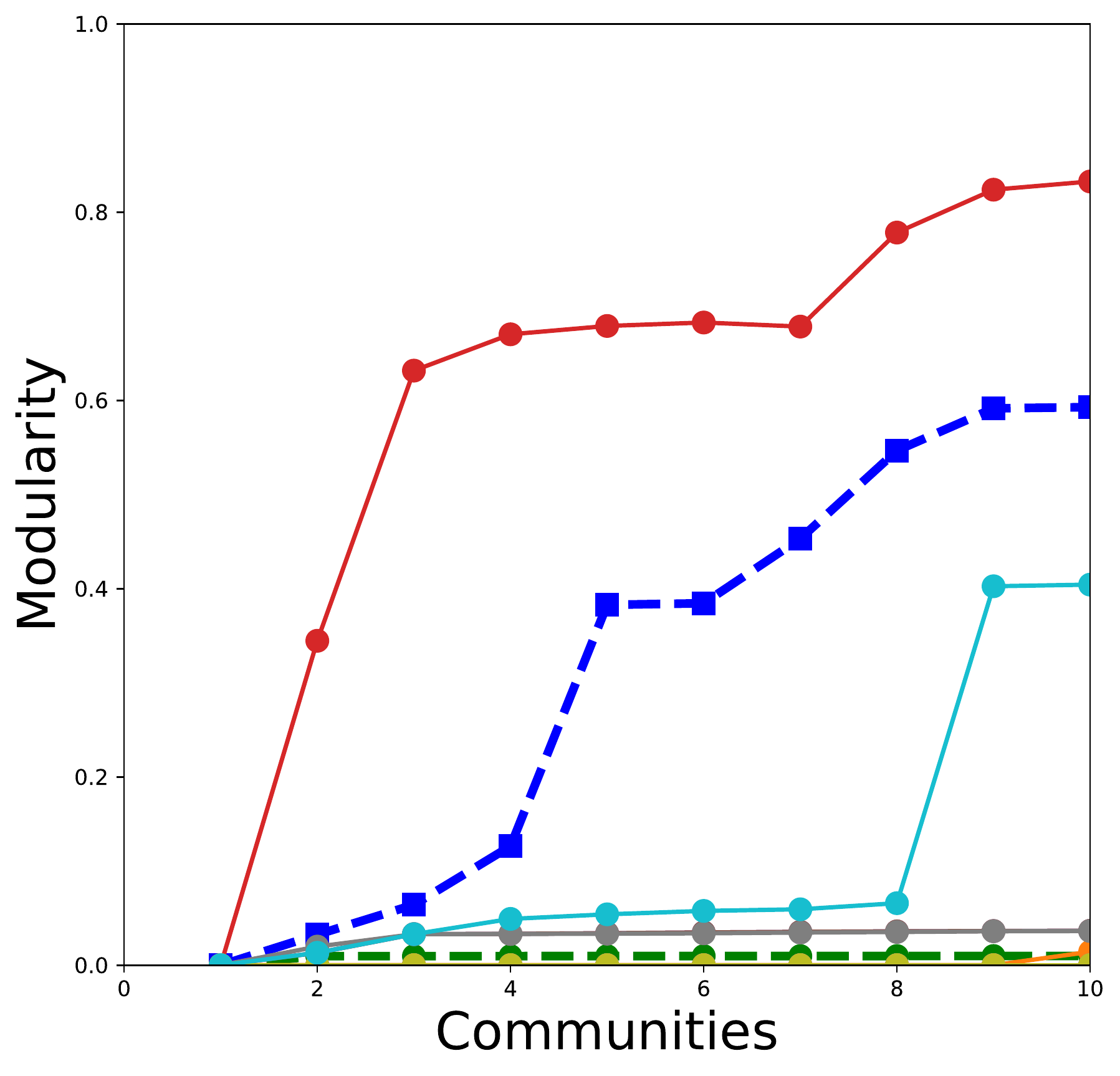}}\hspace{2mm}
		\subfigure[dolphins]{\includegraphics[width=42mm]{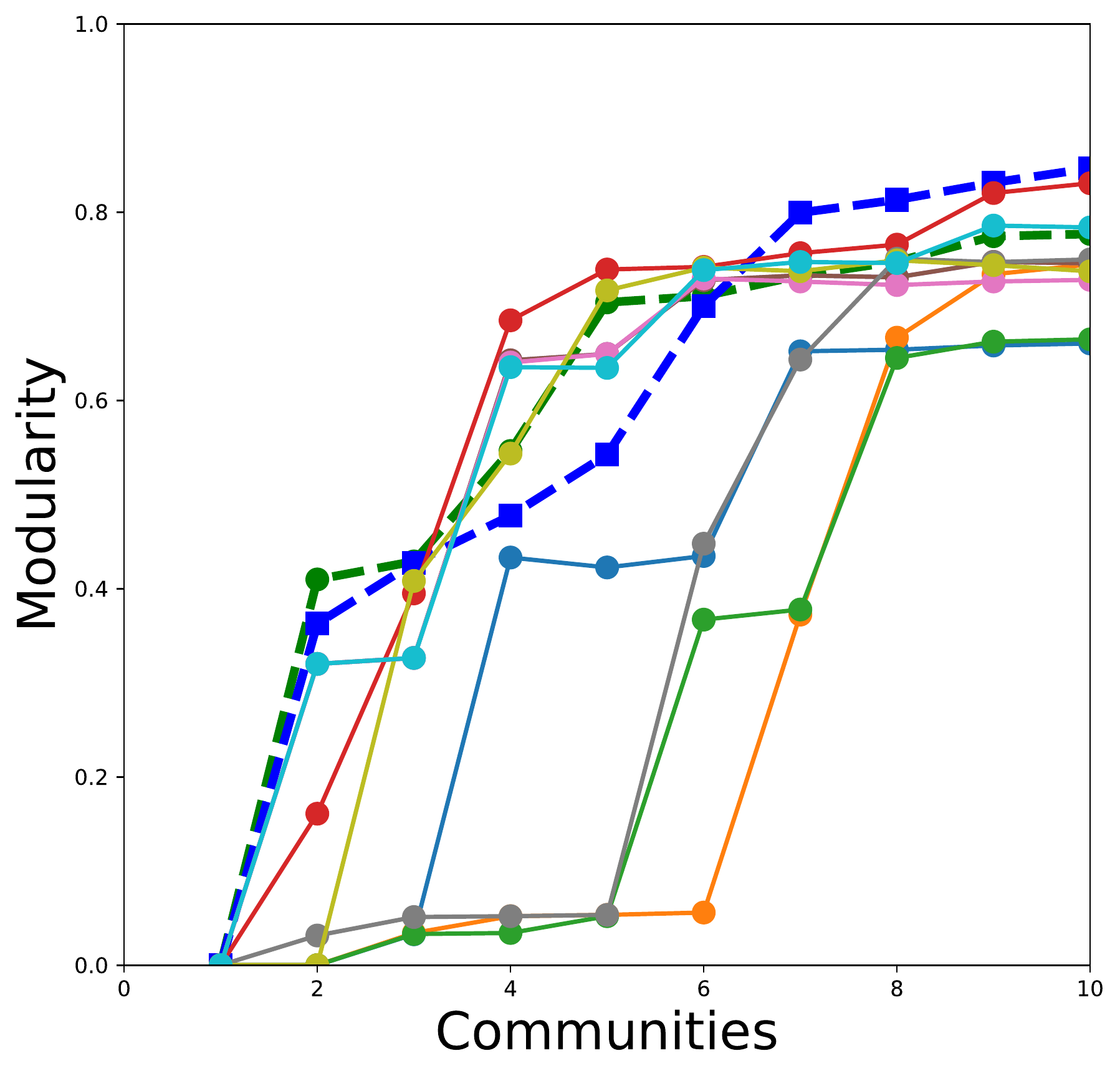}}\vspace{2mm}
		\subfigure[football]{\includegraphics[width=42mm]{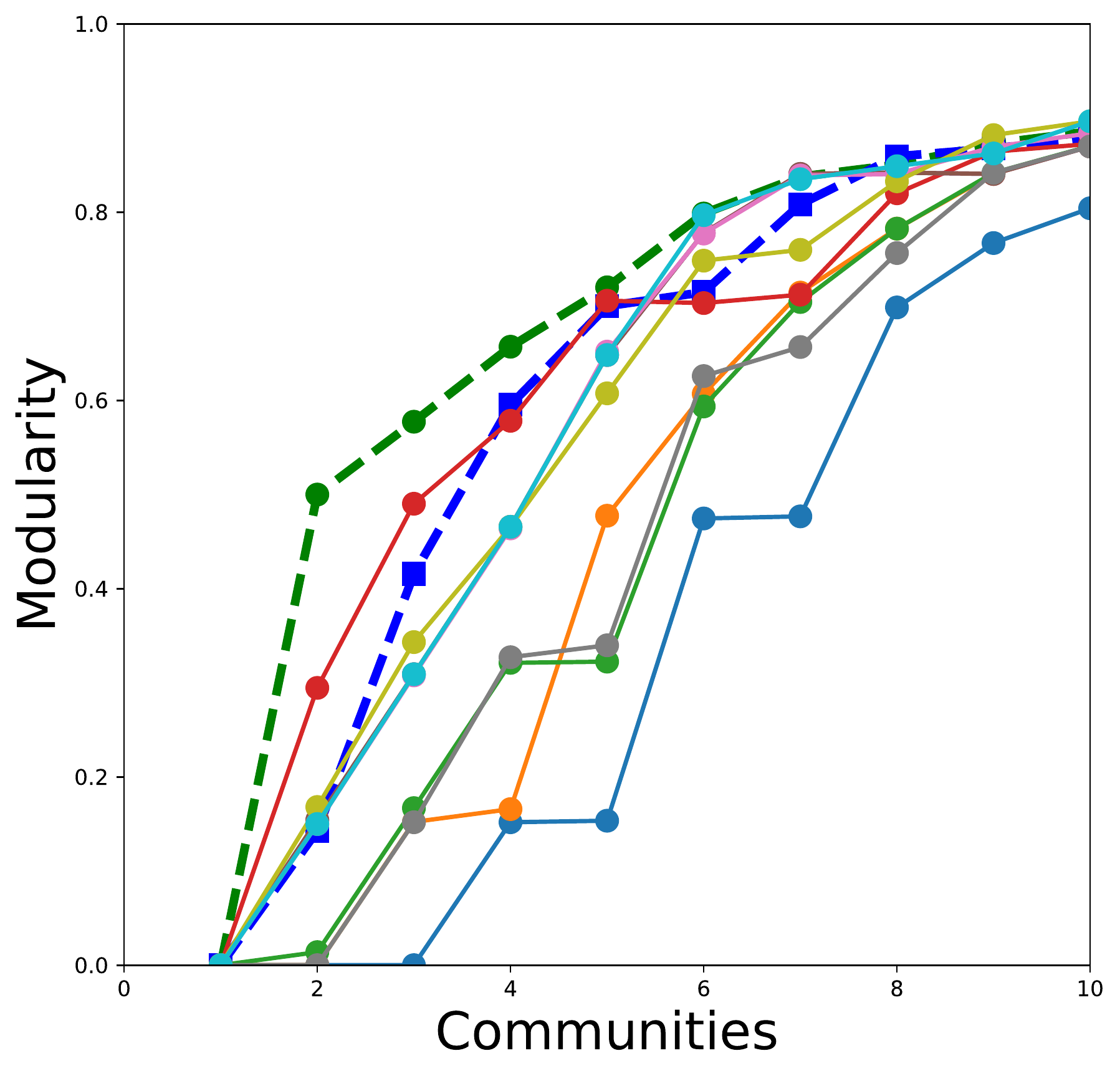}}\hspace{2mm}
		\subfigure[karate]{\includegraphics[width=42mm]{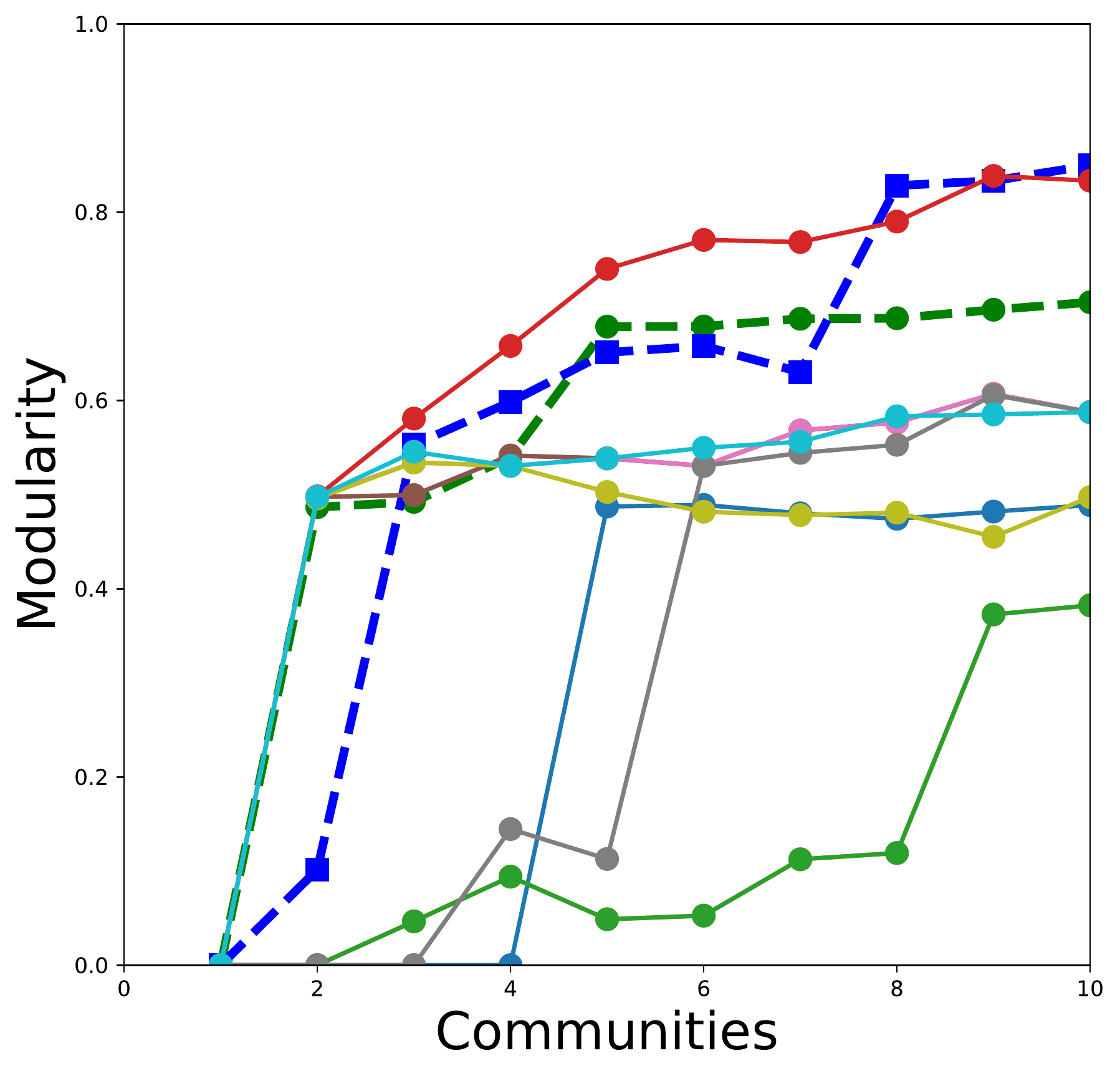}}\vspace{2mm}		\subfigure[lesmis]{\includegraphics[width=42mm]{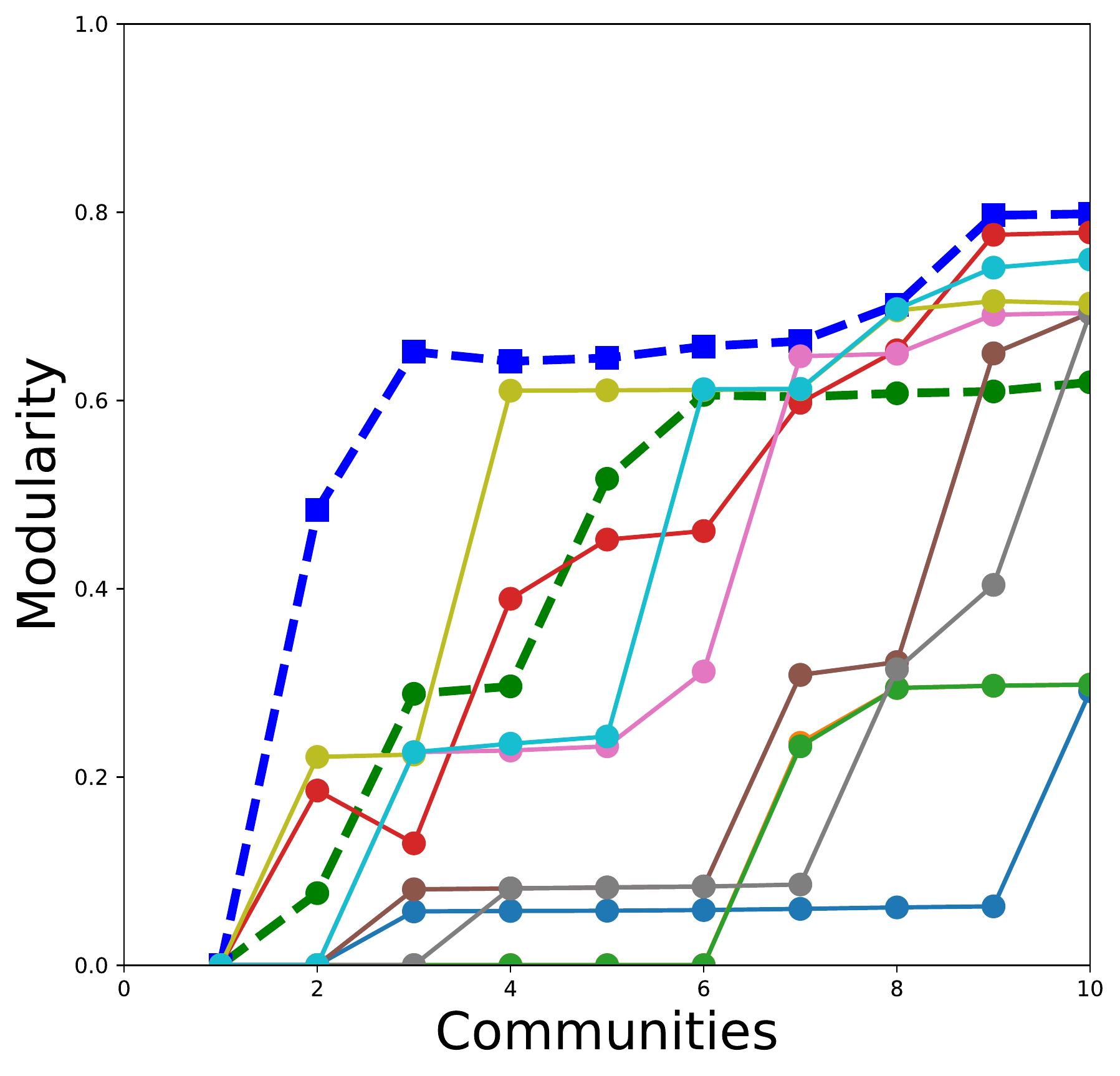}}\hspace{2mm}
		\subfigure[polbooks]{\includegraphics[width=42mm]{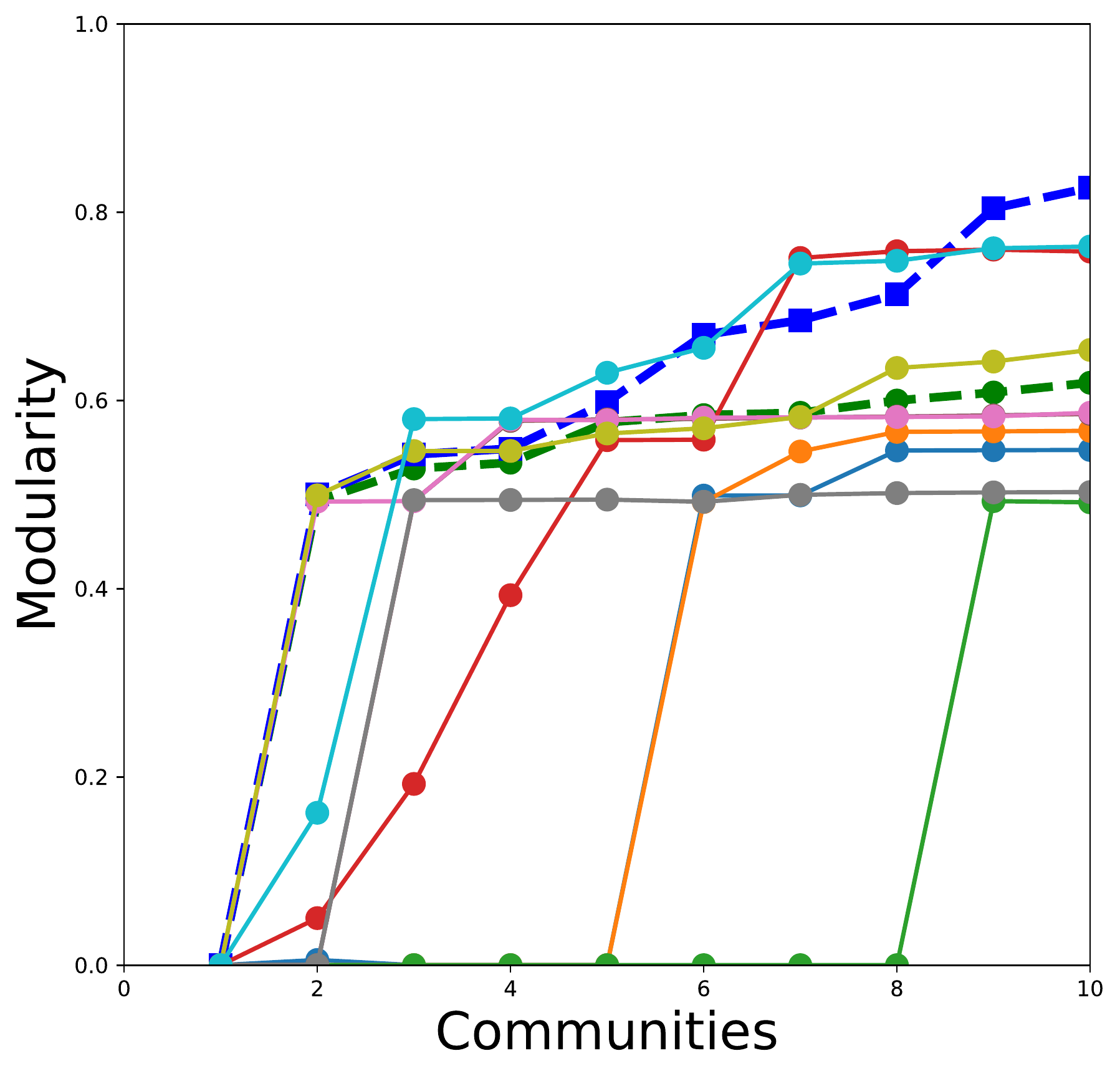}}\vspace{2mm}
		\subfigure{\includegraphics[width=90mm]{figures/legend.pdf}}\vspace{2mm}
		\caption{Modularity results using 12 different criteria for hierarchical community detection on pruned networks.} 
		\label{figura4}
	\end{figure}

	In order to improve the results obtained by hierarchical community detection, we pruned the networks using the strategy discussed above. The results are shown in Figure \ref{figura4}. 
	\\\\
	Comparing Figures \ref{figura2} and \ref{figura4}, we can see that, as intended, noise has been reduced when computing local structural properties and the community detection results have improved in terms of modularity.
	\\
	
	Tables \ref{tabla5} and \ref{tabla6} collect the average modularity obtained by hierarchical community detection with and without network pruning using 10 different local link prediction measures. 
	\\
	\begin{table*}[]
		\centering
		\caption{Average modularity of hierarchical community detection using local link prediction measures, original networks.}
		\begin{tabular}{lrrrrrrrrrrrr}
			\hline
			& \multicolumn{1}{l}{CN} & \multicolumn{1}{l}{AA} & \multicolumn{1}{l}{RA} & \multicolumn{1}{l}{PA} & \multicolumn{1}{l}{JA} & \multicolumn{1}{l}{SO} & \multicolumn{1}{l}{SA} & \multicolumn{1}{l}{HD} & \multicolumn{1}{l}{HP} & \multicolumn{1}{l}{LLHN} & \multicolumn{1}{l}{avg} & \multicolumn{1}{l}{max} \\ \hline
			adjnoun  & 0.00                   & 0.00                   & 0.00                   & 0.49                   & 0.00                   & 0.00                   & 0.00                   & 0.00                   & 0.00                   & 0.00                      & \textbf{0.05}           & \textbf{0.49}            \\ 
			dolphins & 0.01                   & 0.01                   & 0.01                   & 0.53                   & 0.00                   & 0.00                   & 0.01                   & 0.00                   & 0.00                   & 0.01                      & \textbf{0.06}           & \textbf{0.53}            \\ 
			football & 0.47                   & 0.55                   & 0.55                   & 0.50                   & 0.61                   & 0.61                   & 0.60                   & 0.54                   & 0.62                   & 0.59                      & \textbf{0.56}           & \textbf{0.62}            \\ 
			karate   & 0.00                   & 0.00                   & 0.00                   & 0.63                   & 0.38                   & 0.38                   & 0.38                   & 0.10                   & 0.39                   & 0.38                      & \textbf{0.26}           & \textbf{0.63}            \\ 
			lesmis   & 0.00                   & 0.00                   & 0.00                   & 0.53                   & 0.00                   & 0.00                   & 0.00                   & 0.00                   & 0.00                   & 0.01                      & \textbf{0.05}           & \textbf{0.53}            \\ 
			polbook  & 0.00                   & 0.00                   & 0.00                   & 0.43                   & 0.15                   & 0.15                   & 0.43                   & 0.05                   & 0.46                   & 0.54                      & \textbf{0.22}           & \textbf{0.54}            \\ \hline
		\end{tabular}
		\label{tabla5}
	\end{table*}
	
	\begin{table*}[]
		\centering
		\caption{Average modularity of hierarchical community detection using local link prediction measures, pruned networks.}
		\begin{tabular}{lrrrrrrrrrrrr}
			\hline
			& \multicolumn{1}{l}{CN} & \multicolumn{1}{l}{AA} & \multicolumn{1}{l}{RA} & \multicolumn{1}{l}{PA} & \multicolumn{1}{l}{JA} & \multicolumn{1}{l}{SO} & \multicolumn{1}{l}{SA} & \multicolumn{1}{l}{HD} & \multicolumn{1}{l}{HP} & \multicolumn{1}{l}{LLHN} & \multicolumn{1}{l}{avg} & \multicolumn{1}{l}{max} \\ \hline
			adjnoun  & 0.00                   & 0.00                   & 0.00                   & 0.61                   & 0.03                   & 0.03                   & 0.00                   & 0.03                   & 0.00                   & 0.11                      & \textbf{0.08}            & \textbf{0.61}            \\ 
			dolphins & 0.39                   & 0.27                   & 0.28                   & 0.59                   & 0.56                   & 0.56                   & 0.56                   & 0.35                   & 0.54                   & 0.57                      & \textbf{0.47}            & \textbf{0.59}            \\ 
			football & 0.35                   & 0.46                   & 0.46                   & 0.60                   & 0.57                   & 0.57                   & 0.58                   & 0.46                   & 0.57                   & 0.58                      & \textbf{0.52}            & \textbf{0.60}            \\ 
			karate   & 0.29                   & 0.12                   & 0.12                   & 0.65                   & 0.49                   & 0.49                   & 0.50                   & 0.31                   & 0.45                   & 0.50                      & \textbf{0.39}            & \textbf{0.65}            \\ 
			lesmis   & 0.07                   & 0.11                   & 0.11                   & 0.44                   & 0.23                   & 0.23                   & 0.37                   & 0.17                   & 0.50                   & 0.41                      & \textbf{0.26}            & \textbf{0.50}            \\ 
			polbook  & 0.26                   & 0.27                   & 0.10                   & 0.48                   & 0.46                   & 0.46                   & 0.51                   & 0.40                   & 0.52                   & 0.56                      & \textbf{0.40}            & \textbf{0.56}            \\ \hline
		\end{tabular}
		\label{tabla6}
	\end{table*}

	Tables \ref{tabla7}, \ref{tabla8}, and \ref{tabla9} include the results of the statistical tests we performed to verify whether the observed improvements on similarity were statistically significant. These tests compare the original edge-betweenness-based hierarchical community detection with the different hierarchical community detection methods that result from resorting to local link prediction heuristics, using both Student's t-test and Wilcoxon's test.
	\\
	\begin{table*}[]
		\centering
		\caption{Student's t-tests performed to check the differences between the original algorithm, based on edge betweenness, and the algorithms based on local link prediction measures (with network pruning).}
		\begin{tabular}{lrrrrrr}
			\hline
			\textit{\textbf{Student’s t-test}} & \multicolumn{1}{l}{\textbf{adjnoun}} & \multicolumn{1}{l}{\textbf{dolphins}} & \multicolumn{1}{l}{\textbf{football}} & \multicolumn{1}{l}{\textbf{karate}} & \multicolumn{1}{l}{\textbf{lesmis}} & \multicolumn{1}{l}{\textbf{polbook}} \\ \hline
			Bn vs CN2                          & \textbf{0.000109}                     & \textbf{0.000087}                      & \textbf{0.001195}                      & \textbf{0.000001}                    & \textbf{0.000000}                    & \textbf{0.000607}                     \\
			Bn vs AA2                          & \textbf{0.001536}                     & \textbf{0.003542}                      & \textbf{0.008012}                      & \textbf{0.000003}                    & \textbf{0.000003}                    & \textbf{0.002095}                     \\
			Bn vs RA2                          & \textbf{0.007567}                     & \textbf{0.000515}                      & \textbf{0.006797}                      & \textbf{0.000003}                    & \textbf{0.000010}                    & \textbf{0.000015}                     \\
			Bn vs PA2                          & \textbf{0.000000}                     & \textbf{0.005556}                      & \textbf{0.003528}                      & \textbf{0.000000}                    & \textbf{0.004172}                    & \textbf{0.039515}                     \\
			Bn vs JA2                          & \textbf{0.000196}                     & \textbf{0.018640}                      & \textbf{0.034081}                      & \textbf{0.000020}                    & \textbf{0.007429}                    & \textbf{0.042177}                     \\
			Bn vs SO2                          & \textbf{0.000196}                     & \textbf{0.018640}                      & \textbf{0.034081}                      & \textbf{0.000020}                    & \textbf{0.007429}                    & \textbf{0.042177}                     \\
			Bn vs SA2                          & \textbf{0.000033}                     & \textbf{0.010934}                      & \textbf{0.046849}                      & \textbf{0.000029}                    & 0.071101                             & \textbf{0.000239}                     \\
			Bn vs HD2                          & \textbf{0.000006}                     & \textbf{0.008057}                      & \textbf{0.007651}                      & \textbf{0.000017}                    & \textbf{0.001801}                    & \textbf{0.000643}                     \\
			Bn vs HP2                          & \textbf{0.001209}                     & 0.146216                               & \textbf{0.031129}                      & \textbf{0.000005}                    & 0.198531                             & \textbf{0.000649}                     \\
			Bn vs LLHN2                        & \textbf{0.000022}                     & 0.210601                               & 0.070239                               & \textbf{0.000052}                    & 0.690262                             & \textbf{0.000270}                     \\ \hline  
		\end{tabular}
		\label{tabla7}
	\end{table*}
	
	\begin{table*}[]
	\centering
	\caption{Wilcoxon's tests performed to check the differences between the original algorithm, based on edge betweenness, and the algorithms based on local link prediction measures (with network pruning).}
	\begin{tabular}{lrrrrrr}
		\hline
		\textit{\textbf{Wilcoxon’s test}} & \multicolumn{1}{l}{\textbf{adjnoun}} & \multicolumn{1}{l}{\textbf{dolphins}} & \multicolumn{1}{l}{\textbf{football}} & \multicolumn{1}{l}{\textbf{karate}} & \multicolumn{1}{l}{\textbf{lesmis}} & \multicolumn{1}{l}{\textbf{polbook}} \\ \hline
		Bn vs CN2                         & \textbf{0.000625}                     & \textbf{0.000132}                      & \textbf{0.000132}                      & \textbf{0.000182}                    & \textbf{0.000132}                    & \textbf{0.000132}                     \\
		Bn vs AA2                         & \textbf{0.003307}                     & \textbf{0.000132}                      & \textbf{0.000132}                      & \textbf{0.000132}                    & \textbf{0.000132}                    & \textbf{0.000132}                     \\
		Bn vs RA2                         & \textbf{0.019593}                     & \textbf{0.000132}                      & \textbf{0.000155}                      & \textbf{0.000132}                    & \textbf{0.000132}                    & \textbf{0.000132}                     \\
		Bn vs PA2                         & \textbf{0.000132}                     & \textbf{0.003762}                      & \textbf{0.000132}                      & \textbf{0.000155}                    & \textbf{0.019593}                    & \textbf{0.044208}                     \\
		Bn vs JA2                         & \textbf{0.000132}                     & \textbf{0.004274}                      & \textbf{0.000293}                      & \textbf{0.001116}                    & \textbf{0.003762}                    & \textbf{0.000837}                     \\
		Bn vs SO2                         & \textbf{0.000132}                     & \textbf{0.004274}                      & \textbf{0.000293}                      & \textbf{0.001116}                    & \textbf{0.003762}                    & \textbf{0.000837}                     \\
		Bn vs SA2                         & \textbf{0.000342}                     & \textbf{0.004274}                      & \textbf{0.000342}                      & \textbf{0.000724}                    & 0.107466                             & \textbf{0.001116}                     \\
		Bn vs HD2                         & \textbf{0.000132}                     & \textbf{0.000155}                      & \textbf{0.000155}                      & \textbf{0.000182}                    & \textbf{0.000724}                    & \textbf{0.000132}                     \\
		Bn vs HP2                         & \textbf{0.001285}                     & \textbf{0.019593}                      & 0.070157                               & \textbf{0.000625}                    & 0.658009                             & \textbf{0.024224}                     \\
		Bn vs LLHN2                       & \textbf{0.000132}                     & 0.098957                               & 0.227330                               & \textbf{0.000724}                    & 0.277241                             & \textbf{0.002225}                     \\ \hline
	\end{tabular}
	\label{tabla8}
	\end{table*}
	
	\begin{table*}[]
	\centering
	\caption{Statistical tests performed to check the impact of network pruning on the results obtained by hierarchical community detection using different local link prediction heuristics.}
	\begin{tabular}{lrrrrrr}
		\hline
		\textit{\textbf{Student’s t-test}} & \multicolumn{1}{l}{\textbf{adjnoun}} & \multicolumn{1}{l}{\textbf{dolphins}} & \multicolumn{1}{l}{\textbf{football}} & \multicolumn{1}{l}{\textbf{karate}} & \multicolumn{1}{l}{\textbf{lesmis}} & \multicolumn{1}{l}{\textbf{polbook}} \\ \hline
		CN1 vs CN2                         & 0.067889                              & \textbf{0.000196}                      & \textbf{0.001285}                      & \textbf{0.000000}                    & \textbf{0.000196}                    & \textbf{0.000438}                     \\
		AA1 vs AA2                         & \textbf{0.003346}                     & \textbf{0.000196}                      & \textbf{0.021802}                      & \textbf{0.000010}                    & \textbf{0.000982}                    & \textbf{0.000655}                     \\
		RA1 vs RA2                         & \textbf{0.007686}                     & \textbf{0.000196}                      & \textbf{0.040136}                      & \textbf{0.000007}                    & \textbf{0.000982}                    & \textbf{0.002218}                     \\
		PA1 vs PA2                         & 0.136498                              & 0.053406                               & \textbf{0.003307}                      & \textbf{0.035732}                    & \textbf{0.007013}                    & 0.197833                              \\
		JA1 vs JA2                         & \textbf{0.000132}                     & \textbf{0.000132}                      & \textbf{0.001944}                      & 0.285207                             & \textbf{0.000196}                    & 0.077767                              \\
		SO1 vs SO2                         & \textbf{0.000132}                     & \textbf{0.000132}                      & \textbf{0.001944}                      & 0.285207                             & \textbf{0.000196}                    & 0.077767                              \\
		SA1 vs SA2                         & 0.108809                              & \textbf{0.000132}                      & \textbf{0.000967}                      & 0.390730                             & \textbf{0.000196}                    & 0.809204                              \\
		HD1 vs HD2                         & \textbf{0.000132}                     & \textbf{0.000132}                      & \textbf{0.002902}                      & \textbf{0.021373}                    & \textbf{0.000293}                    & \textbf{0.005684}                     \\
		HP1 vs HP2                         & \textbf{0.043114}                     & \textbf{0.000196}                      & \textbf{0.026876}                      & 0.647238                             & \textbf{0.000132}                    & \textbf{0.070157}                     \\
		LLHN1 vs LLHN2                     & \textbf{0.000132}                     & \textbf{0.000132}                      & \textbf{0.021802}                      & 0.083299                             & \textbf{0.000196}                    & 0.872118                              \\ \hline
	\end{tabular}
	\label{tabla9}
	\end{table*}

\section{Conclusion}

In this paper, we have advocated for the use of the local network structural properties employed by local link prediction methods in hierarchical community detection. 
\\\\
The original Girvan and Newman's algorithm, based on edge betweenness, obtains excellent results but is too costly from a computational point of view. Radicchi's alternative proposal, based on an edge clustering coefficient, is more efficient yet fails sometimes. In our experiments, we have shown how local link prediction provides the key to both quality results and efficient algorithms.
\\\\
Given that hierarchical community detection is an iterative greedy algorithm, when basing our decisions on local network properties, we run the risk of taking the wrong decisions due to noise in our local measurements. We have also shown that network pruning can help reduce such noise and improve the results of hierarchical community detection based on local link prediction measures.
\\


	\bibliographystyle{ieeetr}
	\bibliography{bibliography}
	
\end{document}